\documentclass[sigconf]{aamas}

\usepackage{balance} 

\usepackage{todonotes}
\usepackage{algorithm}
\usepackage{algpseudocodex}
\usepackage{subfigure}  
\usepackage{setspace}
\usepackage{multirow} 
\usepackage{natbib}
\usepackage{amsmath}
\usepackage{enumitem}

\def\X{\mathcal{X}}

\def\Z{\mathcal{Z}}
\def\R{\mathcal{R}}

\def\bR{\mathbb{R}}

\def\JS{\mathrm{JS}}
\def\POT{\mathrm{WAE}}

\def\L{\mathcal{L}}

\def\E{\mathbb{E}}

\newtheorem{theorem}{Theorem}

\theoremstyle{remark}

\newtheorem{definition}[theorem]{Definition}

\theoremstyle{definition}

\theoremstyle{remark}

\newcommand{\W}{\mathcal{W}}

\makeatletter
\gdef\@copyrightpermission{
  \begin{minipage}{0.2\columnwidth}
   \href{https://creativecommons.org/licenses/by/4.0/}{\includegraphics[width=0.90\textwidth]{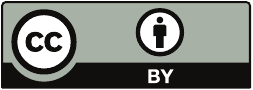}}
  \end{minipage}\hfill
  \begin{minipage}{0.8\columnwidth}
   \href{https://creativecommons.org/licenses/by/4.0/}{This work is licensed under a Creative Commons Attribution International 4.0 License.}
  \end{minipage}
  \vspace{5pt}
}
\makeatother

\setcopyright{ifaamas}
\acmConference[AAMAS '25]{Proc.\@ of the 24th International Conference
on Autonomous Agents and Multiagent Systems (AAMAS 2025)}{May 19 -- 23, 2025}
{Detroit, Michigan, USA}{Y.~Vorobeychik, S.~Das, A.~Nowé  (eds.)}
\copyrightyear{2025}
\acmYear{2025}
\acmDOI{}
\acmPrice{}
\acmISBN{}

\acmSubmissionID{<<262>>}

\title{Imitation from Diverse Behaviors: Wasserstein Quality Diversity Imitation Learning with Single-Step Archive Exploration}

\author{Xingrui Yu }
\authornote{Equal contribution}
\authornote{Corresponding author}
\affiliation{
  \institution{CFAR, IHPC, Agency for Science, Technology and Research}
  \city{}
  \country{Singapore}}
\email{yu\_xingrui@cfar.a-star.edu.sg}

\author{Zhenglin Wan}
\authornotemark[1]
\affiliation{
  \institution{School of Data Science, The Chinese University of Hong Kong, Shenzhen}
  \city{Shenzhen}
  \country{China}}
\email{vanzl3386@gmail.com}

\author{David Mark Bossens}
\affiliation{
  \institution{CFAR, IHPC, Agency for Science, Technology and Research}
  \city{}
  \country{Singapore}}
\email{david\_bossens@cfar.a-star.edu.sg}

\author{Yueming Lyu}
\affiliation{
  \institution{CFAR, IHPC, Agency for Science, Technology and Research}
  \city{}
  \country{Singapore}}
\email{lyu\_yueming@cfar.a-star.edu.sg}

\author{Qing Guo}
\affiliation{
  \institution{CFAR, IHPC, Agency for Science, Technology and Research}
  \city{}
  \country{Singapore}}
\email{guo\_qing@cfar.a-star.edu.sg}

\author{Ivor W. Tsang}
\affiliation{
  \institution{CFAR, IHPC, Agency for Science, Technology and Research}
  \city{}
  \country{Singapore}}
\affiliation{
  \institution{College of Computing and Data Science, NTU}
  \city{}
  \country{Singapore}}
\email{ivor\_tsang@cfar.a-star.edu.sg}

\begin{abstract}
Learning diverse and high-performance behaviors from a limited set of demonstrations is a grand challenge. Traditional imitation learning methods usually fail in this task because most of them are designed to learn one specific behavior even with multiple demonstrations. Therefore, novel techniques for \textit{quality diversity imitation learning}, which bridges the quality diversity optimization and imitation learning methods, are needed to solve the above challenge. This work introduces Wasserstein Quality Diversity Imitation Learning (WQDIL), which 1) improves the stability of imitation learning in the quality diversity setting with latent adversarial training based on a Wasserstein Auto-Encoder (WAE), and 2) mitigates a behavior-overfitting issue using a measure-conditioned reward function with a single-step archive exploration bonus.
Empirically, our method significantly outperforms state-of-the-art IL methods, achieving near-expert or beyond-expert QD performance on the challenging continuous control tasks derived from MuJoCo environments.
\end{abstract}


\newcommand{\BibTeX}{\rm B\kern-.05em{\sc i\kern-.025em b}\kern-.08em\TeX}

\begin{document}


\pagestyle{fancy}
\fancyhead{}


\maketitle 


\section{Introduction}
\label{sect:intro}
Imitation Learning (IL) aims to mimic an expert's behavior by learning from the demonstrations. IL has achieved great success in many real-world applications, such as robotics \citep{manipulation}, autonomous driving \citep{driving}, and drone control \citep{drone}. However, most of the traditional imitation learning methods were designed to learn one specific behavior, even given multiple demonstrations. 

Learning diverse and high-quality policies is the ultimate goal of many real-world applications like robotic locomotion tasks \citep{batra2023proximal}. Recent literature has demonstrated that quality diversity reinforcement learning (QDRL) \citep{batra2023proximal,tjanaka2022approximating} is a promising direction for achieving this goal. Prior methods that combine Differentiable Quality Diversity (DQD) \citep{DQD} with off-policy RL achieved diverse and relatively high-performance policies. However, the performance gap between standard RL and QDRL still exists \citep{batra2023proximal}. More recently, Batra et al. \citep{batra2023proximal} mitigate this gap by leveraging the on-policy RL method PPO \citep{PPO} with DQD. PPO estimates the gradients of diversity and performance objectives from the online collected data. Then, the estimated gradients are used by DQD methods like CMA-MAEGA \citep{CMA-MAEGA} that maintain a single search point and move through the behavior space by filling new regions. The synergy between PPO and DQD results in a state-of-the-art QDRL method, Proximal Policy Gradient Arborescence (PPGA) \citep{batra2023proximal}, that achieves the ultimate goal of robotic locomotion. Generally, this kind of QDRL method finds a diverse archive of high-performing locomotion behaviors for an agent by combining PPO gradient approximations with Differentiable Quality Diversity algorithms.

However, the success of QDRL heavily relies on high-quality reward functions, which can be intractable in practice. Quality Diversity Imitation Learning (QDIL) offers a more flexible strategy for learning diverse and high-quality policies from demonstrations with diverse behaviors. In the literature, adversarial IL methods such as GAIL \citep{ho2016generative} have achieved great success in learning specific behaviors for robotic locomotion tasks. Therefore, a naive solution for QDIL is to apply adversarial IL to estimate rewards from the given demonstrations and then leverage the estimated rewards for learning diverse and high-quality policies. Unfortunately, adversarial IL methods suffer from the \textbf{training instability} issue, which usually results in worse-than-demonstrator performance \citep{ho2016generative}. Moreover, when the demonstrations only contain a few behaviors, the rewards learned by adversarial IL techniques will be \textbf{behavior-overfitted} and unable to guide the agent to learn more diverse behaviors beyond the demonstrations.

\begin{figure*}[htbp]
    \centering
    \includegraphics[width=1.0\linewidth]{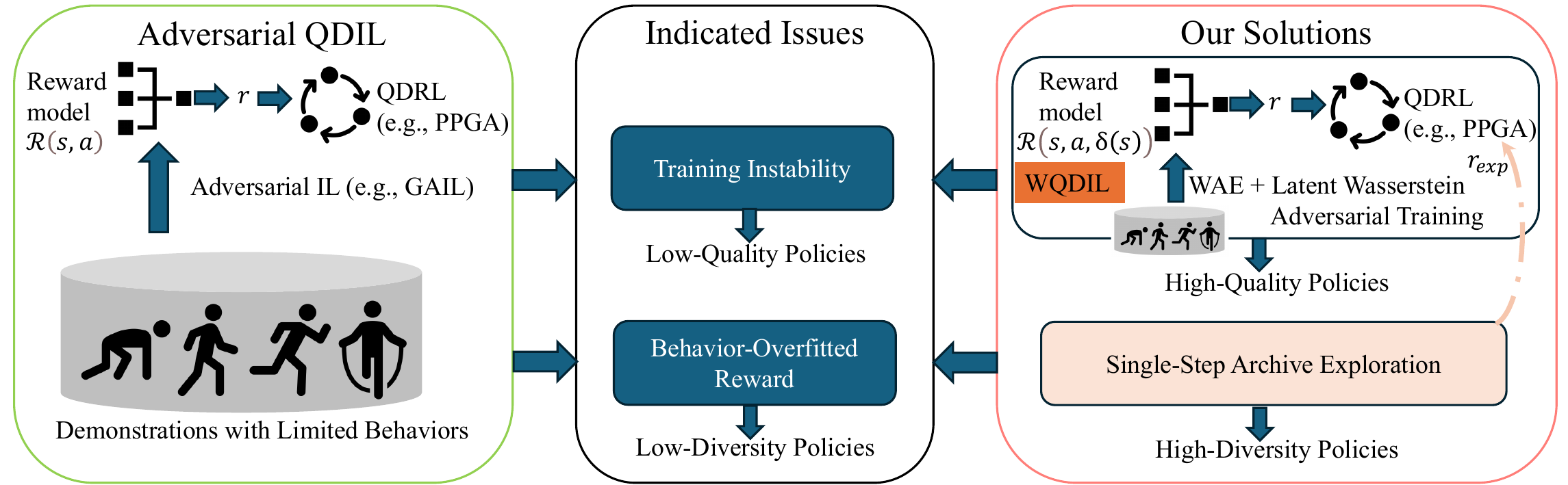}
    \caption{Illustration of the two issues of the Adversarial QDIL (i.e., training instability and behavior-overfitted reward) and their corresponding solutions (i.e. WQDIL and Single-Step Archive Exploration). $\delta(s)$ means the Markovian Measure Proxy of state $s$, a.k.a. the single-step measure.}
    \label{fig:issues_and_solutions}
\end{figure*}

The training instability issue and the behavior-overfitted reward issue heavily limit the adversarial QDIL to learning diverse and high-quality policies with limited demonstrations. In this work, we propose two synergic strategies to overcome these two challenging issues. The first strategy aims to stabilize the training of the reward model, and the second strategy focuses on encouraging the agent to do behavior space exploration. To develop the first strategy, we propose to stabilize the reward learning by applying Wasserstein adversarial training within the latent space of the Wasserstein Auto-Encoder (WAE) \citep{tolstikhin2017wasserstein}. Similar to VAE \cite{KW14}, WAE keeps the good properties of stable training and a nice latent manifold structure while generating higher-quality images than GAN \cite{tolstikhin2017wasserstein}. Therefore, we propose to apply WAE to enable a more stable training of reward model in adversarial QDIL. In addition, we propose latent Wasserstein adversarial training to further improve the consistency of the reward training stability.  
For the second strategy, we introduce a bonus that enables the agent to collect data with more diverse behaviors via single-step archive exploration. We call the resulting method Wasserstein Quality Diversity Imitation Learning (WQDIL) with Single-Step Archive Exploration (SSAE). 
Figure \ref{fig:issues_and_solutions} illustrates the two issues of the Adversarial QDIL (i.e., training instability and behavior-overfitted reward) and the corresponding solutions (i.e. WQDIL and Single-Step Archive Exploration).
The synergy between WQDIL and SSAE adherently address the two issues of adversarial QDIL, resulting in diverse and high-quality policies when learning with limited demonstrations. 

We summarize our contributions as follows:
\begin{itemize}[leftmargin=*] 
\item First, we indicate the two main issues of the naive QDIL solution, i.e., the \textbf{training instability} and the \textbf{behavior-overfitted reward} issues of the adversarial QDIL approach.
\item Second, we propose Wasserstein Quality Diversity Imitation Learning (WQDIL) to address the training instability issue by applying \textbf{Wasserstein adversarial training within the latent space of the Wasserstein Auto-Encoder (WAE)}. 
\item Third, to alleviate the behavior-overfitted reward issue, we design a \textbf{measure-conditioned reward}, which makes the reward function sensitive to the local measure space, together with a \textbf{measure-based bonus}, which encourages the agent to collect data with more diverse behaviors. 
\end{itemize}

\begin{figure}[!h]
    \centering
    \includegraphics[width=1.0\linewidth]{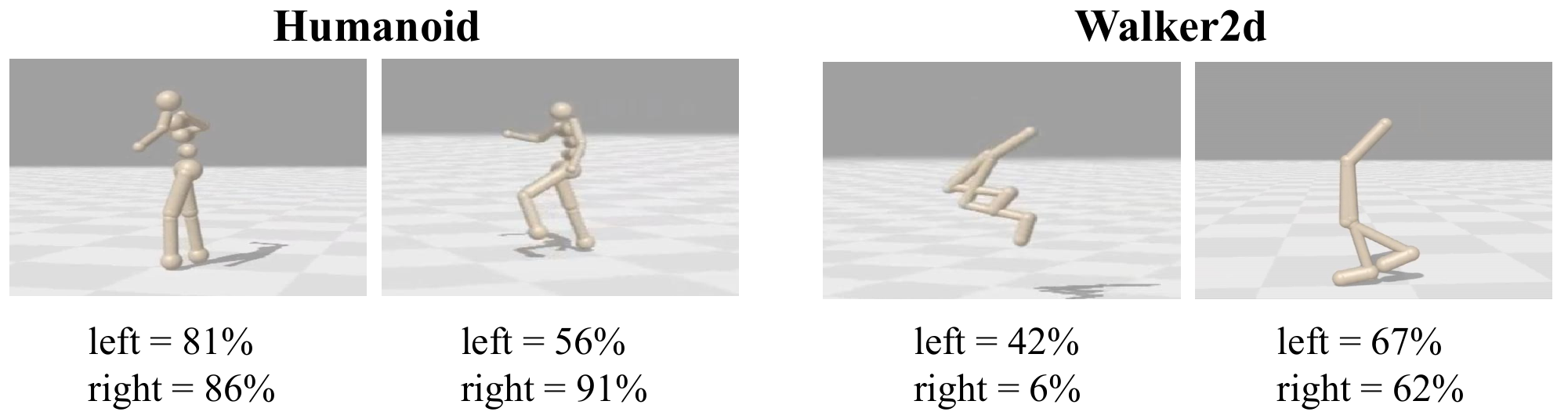}
    \vspace{-0.5cm}
    \caption{Illustration of diverse behaviors learned by our Quality Diversity Imitation Learning framework on Humanoid and Walker2d, where each column represents one behavior. The ``left'' and ```right'' means the proportion of time the left leg or right leg contacting the ground. 
    }
    \label{diverse_behavior}
\end{figure}

\section{Background and Related Work}
This section provides background and related works for Imitation Learning (Section \ref{sect:bg_il}), Quality Diversity Reinforcement Learning (Section \ref{sect:bg_qdrl}), Quality Diversity Imitation Learning (Section \ref{sect:bg_qdil}), Training Stability (Section \ref{sect:bg_stability}), and Exploration (Section \ref{sect:bg_exploration}). 

\subsection{Reinforcement Learning and Imitation Learning}
\label{sect:bg_il}
Reinforcement Learning (RL) searches for policies that maximize cumulative reward in an environment, typically assuming the discrete-time Markov Decision Process (MDP) formalism $(S, A, r, P, \gamma)$. Here, $S$ and $A$ are the state and action spaces, $r(s, a)$ is the reward function, $P(s'|s, a)$ defines state transition probabilities, and $\gamma$ is the discount factor. The traditional RL objective is to maximize the discounted episodic return of a policy $\mathbb{E}\left[\sum_{k=0}^{T-1} \gamma^k r(s_k, a_k)\right]$ where $T$ is the episode length.

Imitation learning (IL) trains an agent to mimic expert behaviors from demonstrations \citep{IL_survey}. Behavior Cloning (BC) uses supervised learning to imitate expert behavior but suffers from severe error accumulation \citep{BC_limitation}. 
Inverse Reinforcement Learning (IRL) seeks to recover a reward function from the demonstrations, and use reinforcement learning (RL) to train a policy that best mimics the expert behaviors \citep{IRL}. Early IRL methods estimate rewards in the principle of maximum entropy \citep{max_ent,max_ent_1,max_ent_2}. Adversarial Inverse Reinforcement Learning (AIRL) \citep{AIRL} learns a robust reward function by training the discriminator via logistic regression to classify expert data from policy data. 

\textbf{Adversarial IL} methods treat IRL as a distribution-matching problem. Generative Adversarial Imitation Learning (GAIL) \citep{ho2016generative} trains a discriminator to differentiate between the state-action distribution of the demonstrations and the state-action distribution induced by the agent's policy, and output a reward to guide policy improvement. More recently, Primal Wasserstein Imitation Learning (PWIL) \citep{PWIL} introduces an offline reward function based on an upper bound of the Wasserstein distance between the expert and agent's state-action distributions, avoiding the instability of adversarial IL methods. This paper includes max-entropy IRL (MaxEntIRL), the classic adversarial IL (i.e., GAIL) and the state-of-the-art PWIL as our baselines in QDIL with limited demonstrations.

\vspace{-0.1cm}
\subsection{Quality Diversity Reinforcement Learning}
\label{sect:bg_qdrl}
\textbf{Quality Diversity Optimization}.
Distinct from traditional optimization which aims to find a single solution to maximize the objective, Quality Diversity (QD) optimization aims to find a set of high-quality and diverse solutions in an \( n \)-dimensional continuous space \( \mathbb{R}^n \). Given an objective function \( f: \mathbb{R}^n \rightarrow \mathbb{R} \) and \( k \)-dimensional measure function \( m: \mathbb{R}^n \rightarrow \mathbb{R}^k \), the goal is to find solutions \( \theta \in \mathbb{R}^n \) for each local region in the behavior space $B = m(\mathbb{R}^n)$.
QD algorithms discretize \( B \)  into \( M \) cells, forming an archive $\mathcal{A}$. Formally, the objective is to find a set of solutions \( \{\theta_i\}_{i=1}^M \) which maximises $f(\theta_i)$ for each $i=1,\dots,M$. Each solution \( \theta_i \) corresponds to a cell in $\mathcal{A}$ via its measure \( m(\theta_i) \), forming an archive of high-quality and diverse solutions \citep{qd_definition,QD_def}.

Previous Quality Diversity optimization methods integrate Evolution Strategies (ES) with MAP-Elites \citep{map-elites}, such as Covariance Matrix Adaptation MAP-Elites (CMA-ME) \citep{CMA-ME}. CMA-ME uses CMA-ES \citep{cma-es} as an ES algorithm generating new solutions inserted into the archive, and uses MAP-Elites to retain the highest-performing solution in each cell. CMA-ES adapts its sampling distribution based on archive improvements from offspring solutions. However, traditional ES faces low sample efficiency, especially for high-dimensional parameters such as neural networks. Differentiable Quality Diversity (DQD) enhances exploration and optimization by leveraging gradient information from both the objective and measure functions. The state-of-the-art DQD algorithm, CMA-MAEGA \citep{CMA-MAEGA}, utilizes CMA-ES to maintain a distribution over coefficient sets \(c_i\), rather than directly generating solutions \( \theta \). The objective function is \( g(\theta) = |c_0| f(\theta) + \sum_{j=1}^{k} c_j m_j(\theta) \), where coefficients \(c_i\) weight the gradients of the objective \(f\) and measures \(m_j\). CMA-MAEGA maintains a search policy and samples multiple coefficient sets after computing gradients of $f$ and $m_i$ to generate branched offspring solutions \( \pi_{\theta_1}, \dots, \pi_{\theta_\lambda} \). The archive improvement from each offspring guides the update of the coefficient distribution, which is then used to refine the search policy.

Built on the QD optimization paradigm, the \textbf{Quality Diversity Reinforcement Learning (QDRL)} problem can be viewed as maximizing $f(\theta) = \mathbb{E}_{\pi_{\theta}}\left[\sum_{k=0}^{T-1} \gamma^k r(s_k, a_k)\right]$ with respect to diverse $\theta$ in a policy archive defined by measure $m$ \citep{QD-robot}. In QDRL, both the objective and measure are non-differentiable, requiring approximations by DQD approaches. The state-of-the-art QDRL algorithm, Proximal Policy Gradient Arborescence (PPGA), employs a vectorized PPO (VPPO) architecture to approximate the gradients of the objective and measure functions \citep{PPGA}. PPGA introduced the Markovian Measure Proxy (MMP), a surrogate measure function that correlates strongly with the original measure and allows gradient approximation via policy gradient by treating it as a reward function. Specifically, MMP decomposes trajectory-based measures $m$ into individual steps by computing:
\begin{equation}
    m_i(\theta) = \frac{1}{T} \sum_{t=0}^{T} \delta_i(s_t),
\end{equation}
where $\delta_i(s_t)$ represents the single-step measure dependent on the state $s_t$. By breaking down the measure in this way, it becomes state-dependent and adheres to the Markov property. Then PPGA uses $k+1$ parallel environments with distinct reward functions -- one for the original reward and $k$ for the surrogate measures. It approximates the gradients of both the objective and the $k$ measure functions by comparing the policy parameters before and after multiple PPO updates. These gradients are then passed to the modified CMA-MAEGA to update the policy archive.

\subsection{Quality Diversity Imitation Learning}
\label{sect:bg_qdil}
\begin{definition}[Quality-Diversity Imitation Learning (QDIL)]
Given expert demonstrations \( \mathcal{D} = \{(s_i, a_i, \delta(s_i))\}_{i=1}^{n} \), where \( s \),\( a \),\( \delta(s)\) represent state and action and the Markovian measure proxy of state $s$. QDIL aims to learn an archive of diverse policies \( \{\pi_{\theta_i}\}_{i=1}^{M} \) that collectively maximizes \( f(\theta) \) (i.e., cumulative reward) \textit{without access to the true reward}. The archive is defined by a \( k \)-dimensional measure function \( m(\theta) \), representing behavior patterns. After dividing the archive into $M$ cells, the objective of QDIL is to find M solutions, each occupying one cell, to maximize:
\begin{equation}
\label{obj}
\max_{\{\theta_i\}} \sum_{i=1}^{M} f(\theta_i).
\end{equation}
\end{definition}

In this paper, we focus on \textbf{Adversarial QDIL}, in which we apply adversarial IL, based on GAIL \cite{goodfellow2014generative} in particular, to estimate rewards from the given demonstrations $\mathcal{D}$. This adversarial training scheme considers the policy as an adversary and an additional neural network, the disciminator $D$, which discriminates between samples of the policy and samples of the demonstrations. The resulting reward model is based on the discriminator, and indicates the likelihood of the parameters based on the demonstration behaviours. This reward model is then used for the objective of QDIL.

\subsection{Improving Stability of Adversarial IL}
\label{sect:bg_stability}

The methodology of adversarial IL comes from Generative Adversarial Networks (GAN) \citep{goodfellow2014generative}. Although adversarial IL methods like GAIL have achieved great success in low-dimensional environments, they also inherit the training instability issue of GAN that limits its generalization to more diverse tasks \citep{luo2024c}. This problem will be exacerbated in the QDIL setting due to the variety of policies being trained.

Wasserstein GAN (WGAN) \citep{arjovsky2017wasserstein} was proposed to stabilize the training of GANs by minimizing the Wasserstein distance instead of the Jensen-Shannon divergence between true data distribution and generated data distribution. However, the original WGAN with weight clipping can lead to undesired behaviors \citep{gulrajani2017improved}. Gradient penalty (GP) was later proposed to address this issue, and has become a widely-used strategy for improving the stability of GAN and WGAN \citep{roth2017stabilizing,petzka2017regularization,gulrajani2017improved,mescheder2018training,thanh2019improving,qi2020loss}. 
We have applied these stabilization methods in adversarial QDIL, and conduct preliminary investigation experiments on continuous control tasks derived from MuJoCo. However, we only observed negligible effects of these methods on stabilizing adversarial QDIL. 

\textbf{Wasserstein Auto-Encoder (WAE)} \citep{tolstikhin2017wasserstein} minimizes the optimal transport cost $W_c(P_X, P_G)$ based on the novel auto-encoder formulation.
In the resulting optimization problem, the decoder tries to accurately reconstruct the encoded training examples as measured by the cost function $f_c$.
Similar to VAE \citep{KW14}, the encoder tries to simultaneously achieve two conflicting goals: it tries to match the encoded distribution of training examples $Q_Z:=\E_{P_X}[Q(Z|X)]$ to the prior $P_Z$ as measured by any specified divergence $D_z(Q_Z, P_Z)$, while making sure that the latent codes provided to the decoder are informative enough to reconstruct the encoded training examples. Using the squared cost $f_c(x,y)=\|x-y\|_2^2$, WAE keeps the good properties of VAEs (stable training and a nice latent manifold structure) while generating better-quality images than GAN \citep{tolstikhin2017wasserstein}. This observation inspired us to apply WAE to improve the stability of adversarial QDIL. The resulting framework is called Wasserstein Quality Diversity Imitation Learning (WQDIL). 

\subsection{Encouraging exploration in RL and IL}
\label{sect:bg_exploration}
Exploration bonuses have a long history in bandits and reinforcement learning as part of the ``optimism in face of uncertainty'' principle, and include but are not limited to visitation count based approaches. In bandits, upper confidence bound (UCB, e.g. \cite{Auer2002}) computes a bonus based on $\sqrt{\frac{2\ln(t)}{n(a_t)}}$ for action $a_t$ at time $t$. In tabular reinforcement learning, upper confidence reinforcement learning (UCRL) techniques (e.g. \cite{Jaksch2010}) provide upper confidence bounds to the reward based on a bonus proportional to $\sqrt{\frac{1}{n(s_t,a_t)}}$. R-MAX \cite{Brafman2002} and E3 \cite{Kearns2002a} explicitly distinguish between known and unknown states, where in unknown states R-MAX assigns the maximal reward of the MDP and E3 performs balanced wandering (i.e. taking the action with lowest visitation count for the current state). Related concepts have also been applied in deep reinforcement learning \cite{Tang2017,Henaff2022,Machado2020,Seo2021}. Related to exploration bonuses, recent techniques in imitation learning also considered curiosity-based exploration \citep{pathak2017curiosity,burda2018large,giril}, which relates to discrepancy between the transition model's prediction and the true transition. For instance, GIRIL \citep{giril} computes rewards offline by pretraining a reward model using a conditional VAE \citep{sohn2015learning}.

Our measure bonus differs from these approaches in that it is designed to explore the behavior space rather than the (state-)action space. Further, contrasting to UCRL type bonuses, due to dividing by the proportion of visits, rather than the count, our measure bonus does not shrink to zero such that underexplored behaviors will continue to receive a bonus. This is somewhat comparable in aim to the balanced wandering of E3 but differs in that it is a reward bonus rather than a policy definition (since we do not have control over the measures) and that it is maintained throughout the entire optimization process as we are interested in behavioral diversity.

\section{Wasserstein Quality Diversity Imitation Learning (WQDIL)}
\label{sect:wqdil}
This section introduces our proposed framework-WQDIL. We first introduce the behavior overfitted reward problem and propose two solutions called Single-Step Archive Exploration Bonus and Measure Conditioning (Section \ref{sect:ssae}). We then introduce Wasserstein Auto-encoders (Section \ref{sect: WAE}), and propose a WQDIL instance WAE-WGAIL that applies latent Wasserstein adversarial training for WAE optimization (Section \ref{sect:wae4qdil}).

\subsection{Mitigating behavior overfitting}
\label{sect:ssae}
Compared to the vast behavior space, expert behaviors tend to be limited in diversity. This poses a problem for learning diverse behaviors, as the reward model in the traditional IL paradigm focuses solely on shaping rewards around specific expert behaviors while neglecting other behavior patterns. This presents a challenge for balancing exploration and exploitation and makes reward functions local in that they only count a single behavior as high-performing \cite{wan2024qualitydiversityimitationlearning}. We refer to this as the ``behavior-overfitted reward'' problem. This problem significantly inhibits the learning of diverse behaviors, since any RL method inherently converges to high-reward behaviors. To address this problem, we introduce a single-step archive-exploration reward bonus and measure-conditioning, which together ensure exploration across the behavior space as well as the sensitivity of the reward function to the local measure information.

\paragraph{Single-Step Exploration Bonus and Measure Conditioning}
In both the bonus and the measure conditioning, we are primarily interested in single-step measures building on the Markovian Measure Proxy mentioned in Section \ref{sect:bg_qdrl} to make the measure information more fine-grained. This is not merely for convenience purposes but also as this will yield richer data from the demonstrations, which would otherwise be sparse (e.g. only a few episodic measures since we focus on limited-expert-demonstrations in our setting). Meanwhile, focusing on single step measure will allow exploring behaviors with similar episodic measure but different single-step measure for potentially yielding higher fitness within a particular behavioral region. This is because the episodic measure is the average (or sum) of each single-step measure and is less fine-grained.

For the exploration bonus, we establish the single-step archive $\mathcal{A}_{single}$, which corresponds to the state-dependent measure $\delta(s)$. Similar to the behavior archive $\mathcal{A}$, we partition $\mathcal{A}_{single}$ into numerous cells for discretization. Notably, instead of merely recording whether a cell is occupied, we track the \textbf{visitation count} $n_i$ for each cell $C_i$ in $\mathcal{A}_{single}$. The exploration reward bonus is formulated as
\begin{equation}
    r_{exp}(s,a,\delta(s)) = \frac{1}{1 + p(\delta(s))},
\label{eq: r_exp_1}
\end{equation}
where
\begin{equation}
    p(\delta(s)) = \frac{n(C(\delta(s)))}{\sum_i n(C_i)}.
    \label{eq: r_exp_2}
\end{equation}
In these equations, $C_i$ denotes the $i$-th cell in $\mathcal{A}_{single}$, $C(\delta(s))$ represents the cell corresponding to the single-step measure proxy $\delta(s)$, and $n$ signifies the visitation count. Each time a state $s$ activates a cell in $\mathcal{A}_{single}$, the visitation count of that cell is incremented by one. This mechanism allows the single-step archive $\mathcal{A}_{single}$ to be dynamically updated during training.

The exploration bonus assigns higher rewards to regions in $\mathcal{A}_{single}$ that are less frequently visited, thereby promoting the agent to explore unseen behavior patterns. Additionally, once a region within the single-step behavior space has been sufficiently explored, the bonus decreases, facilitating the exploitation of that region to discover high-performing policies. However, note that the bonus is defined relative to the exploration of other measures such that the bonus never shrinks to zero for a particular measure. The form in Eq.~\ref{eq: r_exp_1} also avoids extreme values due to the +1 offset in the denominator. With these features together, the reward bonus can effectively mitigate the ``behavior-overfitted reward'' issue by always encouraging new behavior patterns, thus facilitating diverse behaviors. 

In addition to the exploration bonus, the reward model is also conditioned on the measure. Specifically, using GAIL for imitation learning, the reward model is formulated as
\begin{align}
\label{eq: GAIL}
\tilde{\R}(s,a,\delta(s)) &  =  \max_{\pi} \min_{D} \mathbb{E}_{(s,a) \sim \mathcal{D}} [-\log D(s,a,\delta(s))]  \nonumber \\ & + \mathbb{E}_{(s,a)\sim \pi}[-\log(1-D(s,a,\delta(s)))] .
\end{align}
This contributes to a solution to behavior overfitting: since the goal is to form behaviorally diverse policies, different behaviors require different policies to be imitated -- and hence correspond to a different reward function. Combining with the exploration bonus, we obtain the combined reward model
\begin{equation}
\R(s,a,\delta(s)) =  \tilde{\R}(s,a,\delta(s))   +  r_{exp}(s,a,\delta(s)). 
\end{equation}

\subsection{Wasserstein auto-encoders}
\label{sect: WAE}
The Wasserstein distance is a formulation that occurs within the context of optimal transport \cite{V03}. Kantorovich's formulation of the problem is given by
\begin{equation}
\label{eq:ot}
\W_c(P_x,P_G):=\inf_{\Gamma\in \mathcal{P}(x\sim P_X,y\sim P_G)} \E_{(x,y)\sim\Gamma}[f_c(x,y)]\,,
\end{equation}
where $f_c(x,y)\colon \X \times \X \to \bR_+$ is the cost function
and $\mathcal{P}(x,y)$ is the set of joint distributions of $(x,y)$ with marginals $P_x$ and $P_G$ respectively. We are interested in the set $\X = S \times A \times M$, where $M$ is the measure space.

Auto-encoders aim to represent input data in a low-dimensional latent space. Modern generative models like variational auto-encoders (VAEs) \cite{KW14} and generative adversarial networks (GANs) \cite{goodfellow2014generative} do so by minimizing a discrepancy measure between the data distribution $P_x$ and the generative model $P_G$. To formulate a Wasserstein based generative model, the form in Eq.~\ref{eq:ot} is intractable; however, following \cite{BGT+17}, one can equate the Wasserstein distance to the following WAE objective \cite{tolstikhin2017wasserstein}:
\begin{equation}
\label{eq:wae-obj}
    \begin{aligned}
    D_\POT(P_x,P_G):=&
    \inf_{Q(z \vert x)\in\mathcal{Q}} \E_{P_x} \E_{Q(z \vert x)} \bigl[f_c\bigl(x,P_G(z)\bigr)\bigr] \\&+ \lambda \cdot D_z(Q_z, P_z),\\
    \end{aligned}
\end{equation}
where $P_z$ is the latent distribution, $D_z$ is a divergence measure, and $P_G$ is the decoder (i.e. generative model). We restrict our cost function to the squared Euclidian distance, i.e. $f_c(x,y) = \vert\vert x - y \vert\vert^2_2$, resulting in the $2$-Wasserstein distance.

\textbf{WAE-GAN} \citep{tolstikhin2017wasserstein} is an instance of WAE when choosing $D_z(Q_z, P_z) = D_\JS(Q_z, P_z)$ and using adversarial training to estimate $D_z$. Specifically, WAE-GAN uses an adversary and discriminator in the latent space $\Z$ trying to separate ``true'' points sampled from $P_z$ and ``fake'' ones sampled from $Q_z$ \cite{goodfellow2014generative}. In the imitation learning setting, $P_z$ corresponds to the distribution of latent data obtained from the encoded demonstrations while $Q_z$ corresponds to the distribution of latent data obtained from the encoded trajectory data from the policy.

It has been noted previously that WAE-GAN still suffers from instabilities due to the mode collapse property of GANs \cite{kolouri2018sliced}. Therefore, we propose \textbf{WAE-WGAN}, which is equivalent to WAE-GAN except that it sets the divergence measure to the $1$-Wasserstein distance, i.e. $D_z(Q_z, P_z) = \W_1(Q_z, P_z)$. We choose this option based on results on the improved stability during adversarial training \cite{arjovsky2017wasserstein}. Due to strong duality with $p=1$, the Kantorovic-Rubinstein duality can be applied, i.e. 
\begin{equation}
\W_1(Q_z, P_z) = \sup_{\vert\vert f \vert\vert_{L} \leq 1} \E_{z \sim Q_z}[f(z)] - \E_{z \sim P_z}[f(z)] ,
\end{equation}
where $f$ is a $1$-Lipschitz function $f: \X \to \bR$.

\subsection{WQDIL: WAE Meets Adversarial QDIL}
\label{sect:wae4qdil}
This section introduces our Wasserstein Quality Diversity Imitation Learning (WQDIL) framework, which is built on top of the Proximal Policy Gradient Arborescence (PPGA) algorithm \citep{PPGA}. The pseudo-code of WQDIL is given in Algorithm \ref{alg:wqdil}, where the blue color indicates the WQDIL's operations that are distinct from the PPGA algorithm.

\begin{algorithm}[!h]
\caption{Wasserstein Quality Diversity Imitation Learning (WQDIL)}
\label{alg:wqdil}
\begin{algorithmic}[1]
\State \textbf{Input:} Initial policy $\theta_0$, VPPO instance to approximate $\nabla f$, $\nabla m$ and move the search policy, 
number of QD iterations $N_Q$, number of VPPO iterations to estimate the objective-measure functions and gradients $N_1$, 
number of VPPO iterations to move the search policy $N_2$, branching population size $\lambda$, and an initial step size for xNES $\sigma_g$. \textcolor{blue}{Initial reward model $\mathcal{R}$, Expert demonstrations $\mathcal{D}$}.
\State Initialize the search policy $\theta_\mu = \theta_0$. Initialize NES parameters $\mu, \Sigma = \sigma_g I$
\For{$\text{iter} \gets 1$ to $N$}
    \State \textcolor{blue}{$f, \nabla f, \mathbf{m}, \nabla \mathbf{m} \gets \text{VPPO.compute\_jacobian}(\theta_\mu, \mathcal{R}, \mathbf{m}(\cdot), N_1)$} \Comment{approx gradients $\nabla f$ and $\nabla m$ using the rewards estimated by $\mathcal{R}$ (Algorithm \ref{alg:reward_estimation})}
    \State $\nabla f \gets \text{normalize}(\nabla f), \quad \nabla \mathbf{m} \gets \text{normalize}(\nabla \mathbf{m})$
    \State $\_$ $\gets \text{update\_archive}(\theta_\mu, f, \mathbf{m})$
    \For{$i \gets 1$ to $\lambda$} // branching solutions
        \State $c \sim \mathcal{N}(\mu, \Sigma)$ // sample gradient coefficients
        \State $\nabla_i \gets c_0 \nabla f + \sum_{j=1}^k c_j \nabla m_j$
        \State $\theta_i' \gets \theta_\mu + \nabla_i$
        \State $f', *, \mathbf{m}', * \gets \text{rollout}(\theta_i',\mathcal{R})$
        \State $\Delta_i \gets \text{update\_archive}(\theta_i', f', \mathbf{m}')$ \Comment{get archive improvement of each solution using Algorithm 2 of PPGA paper.}
    \EndFor
    \State Rank gradient coefficients $\nabla_i$ by archive improvement $\Delta_i$
    \State Adapt xNES parameters $\mu = \mu', \Sigma = \Sigma'$ based on improvement ranking $\Delta_i$
    \State $f'(\theta_\mu) = c_{\mu,0} f + \sum_{j=1}^k c_{\mu,j} m_j$, where $c_{\mu} = \mu'$
    \State \textcolor{blue}{$\theta_\mu' = \text{VPPO.train}(\theta_\mu, f', \mathbf{m'}, N_2, \mathcal{R})$ }\Comment{walk search policy using reward model $\mathcal{R}$}
    \State \textcolor{blue}{$\mathcal{R}.\text{update}(\mathcal{D},\theta_\mu')$}  \Comment{update reward model using Algorithm \ref{alg:WAE-GAIL} (Appendix \ref{sect:algorithms}) for WAE-GAIL, Algorithm \ref{alg:WAE-WGAIL} (Appendix \ref{sect:algorithms}) for WAE-WGAIL, or Algorithm \ref{alg:mCWAE-WGAIL} for mCWAE-WGAIL.}
    \If{there is no change in the archive}
        \State Restart xNES with $\mu = 0, \Sigma = \sigma_g I$
        \State Set $\theta_\mu$ to a randomly selected existing cell $\theta_i$ from the archive
    \EndIf
\EndFor
\State \textbf{Output:} Updated policy archive. 
\end{algorithmic}
\end{algorithm}

\begin{algorithm}[htbp]
\caption{Reward Estimation} 
\label{alg:reward_estimation}
\begin{algorithmic}[1]
\State \textbf{Initialize:} Reward model $\mathcal{R}$
\State \textbf{Method: Reward estimation in Algorithm \ref{alg:wqdil} (step 4)} 
\State \texttt{def get\_episode\_reward}(episode, current archive $\mathcal{A}$): 
\State \quad $s_1,a_1,\delta(s_1),s_2,a_2,\delta(s_2)\dots,s_k,a_k,\delta(s_k) \gets \texttt{episode}()$
\State \quad $r_1,r_2,\dots,r_k \gets \tilde{\R}(s,a,\delta(s))$ \Comment{batch reward (see Eq.\ref{eq: GAIL})}
\State \quad compute exploration bonuses $r_{1,exp}, r_{2,exp},\dots,r_{k,exp}$ \Comment{(see Eq.\ref{eq: r_exp_1})}
\State \quad \textbf{For} $i=1,\dots,k$
\State \quad \quad $r_{i} \gets r_i + r_{i,exp}$\Comment{calculate total reward for each step}
\State \quad \textbf{return} $r_1,r_2,\dots,r_k$
\end{algorithmic}
\end{algorithm}

In WQDIL, we first initialize the reward model $\mathcal{R}$ and input the expert demonstrations $\mathcal{D}$. Then we repeat iterations as follows. First, approximate the gradients $\nabla f$ and $\nabla m$ (in l.4) using the rewards estimated from $\mathcal{R}$ (Algorithm \ref{alg:reward_estimation}). Lines 5-15 are QD optimization options used in PPGA. In line 16, we update the search policy using the reward model $\mathcal{R}$. Line 17 updates the reward model $\mathcal{R}$ using the input expert demonstrations $\mathcal{D}$. We now discuss different algorithms for updating the reward model $\mathcal{R}$ in WQDIL.  

\textbf{WAE-GAIL.} 
The first reward updating algorithm is WAE-GAIL, which is a naive adaptation of WAE-GAN in the QDIL framework. Algorithm \ref{alg:WAE-GAIL} in Appendix \ref{sect:algorithms} summarizes the details of WAE-GAIL. 

\textbf{WAE-WGAIL.} Although WAE-GAIL achieves better stability and higher QD performance than GAIL in some tasks, the improvements are not consistent. To further improve WAE-GAIL, we propose to apply Wasserstein adversarial training in the latent space of WAE, which is analogous to the WAE-WGAN proposed in Section~\ref{sect: WAE}. The loss function of the auto-encoder in WAE-WGAIL is based on the reconstruction loss for expert demonstrations, the reconstruction loss for policy data, and the Wasserstein distance between latent variables of expert demonstrations and policy data: 
\begin{align}
\label{eq:wae-wgail loss}
\L(\phi,\psi) &= \L_{\text{recon}}^{e} + \L_{\text{recon}}^{\pi} + \lambda D_z(z^e,z^{\pi})  \nonumber \\
              &= \E_{x^e \sim \mathcal{D},z^e \sim Q_{\phi}(z^e \vert x^e)}[f_c(x^e, G_{\psi}(z^e))] \nonumber \\
              &+ \E_{x^\pi \sim \pi,z^\pi \sim Q_{\phi}(z^\pi \vert x^\pi)}[f_c(x^\pi, G_{\psi}(z^\pi) )]   \nonumber \\
             & + \lambda \E_{z^e,z^{\pi}}\left[D_{\W}(z^e) - D_{\W}(z^{\pi}) \right] ,
\end{align}
where $\phi$ and $\psi$ are the parameters of the encoder and decoder, respectively, $x = (s,a)$ is the state-action pair, $Q_{\phi}$ is the encoder, $G_{\psi}$ is the decoder, and $D_{\W}$ is the Wasserstein discriminator. Algorithm \ref{alg:WAE-WGAIL} in Appendix \ref{sect:algorithms} summarizes the details of WAE-WGAIL. 

\textbf{mCWAE-WGAIL.} Compared to GAIL, WAE-WGAIL has significantly better stability and consistency across different tasks. However, the latent space of WAE-WGAIL is learned from the state-action pairs. Since the demonstrations only contain very limited behaviors, the learned latent space is not adaptive to more diverse behaviors that the agent may encounter. To improve the adaptivity of the latent space and the reward model $\mathcal{R}$, we propose measure-conditioned WAE-WGAIL (mCWAE-WGAIL) that learns latent space by reconstructing state-action-measure triplet, i.e. formulating $x=(s,a,\delta(s))$ in Eq.~\ref{eq:wae-wgail loss}. 
Algorithm \ref{alg:mCWAE-WGAIL} summarizes the details of mCWAE-WGAIL, 
our proposed algorithm for WQDIL.

\begin{figure}[!h]
    \subfigure[HalfCheetah]{\includegraphics[width=0.155\textwidth]{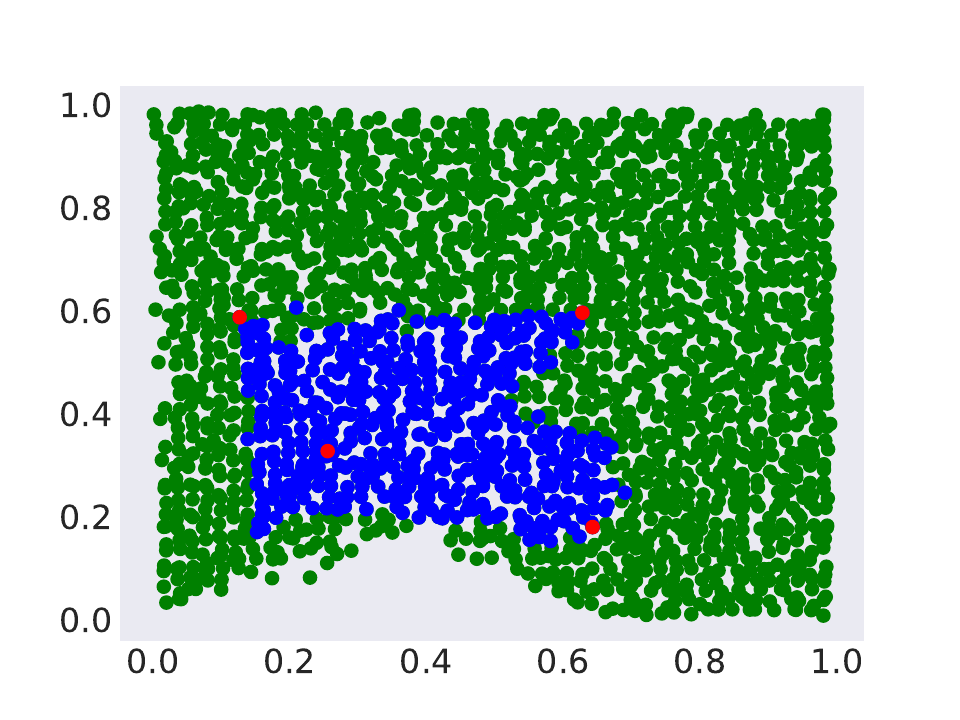}}
    \subfigure[Walker2d]{\includegraphics[width=0.155\textwidth]{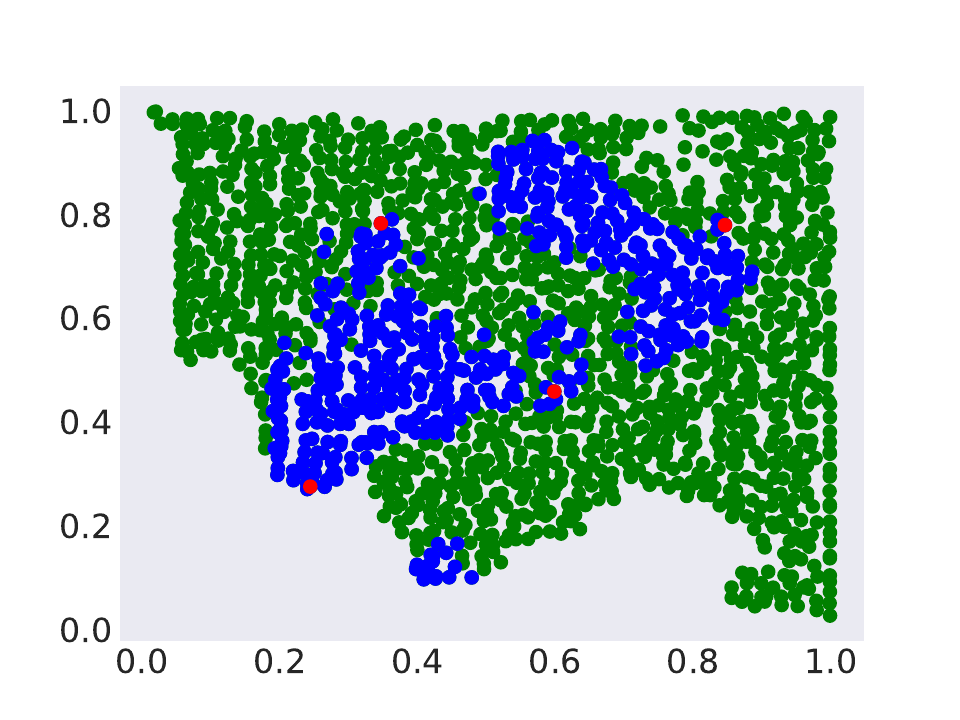}} 
    \subfigure[Humanoid]{\includegraphics[width=0.155\textwidth]{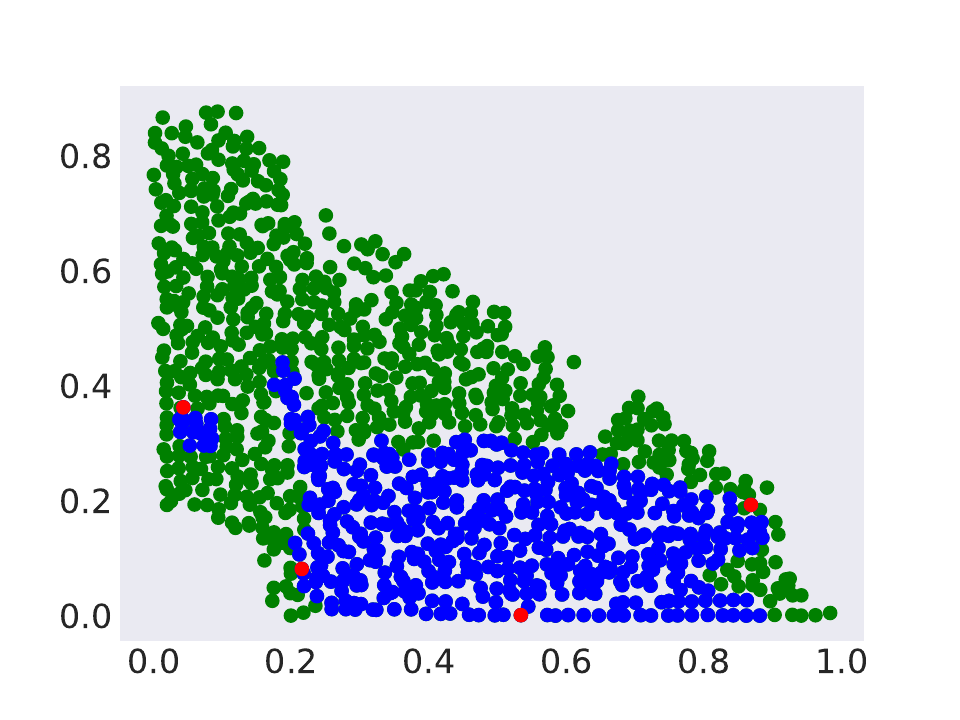}} 
    \vspace{-0.4cm}
    \caption{Visualization of the demonstrations obtained from PPGA archives. The x and y axes are the proportions of time leg 1 and 2, respectively, touch the ground. Green indicates the full expert behavior space, blue indicates the selected top-500 elites, and red indicates the selected demonstrators. }
    \label{fig:top500elites}
\end{figure}

\label{sect:exp_perf}
\begin{figure*}[!h]
    \centering
    \includegraphics[width=1.0\textwidth]{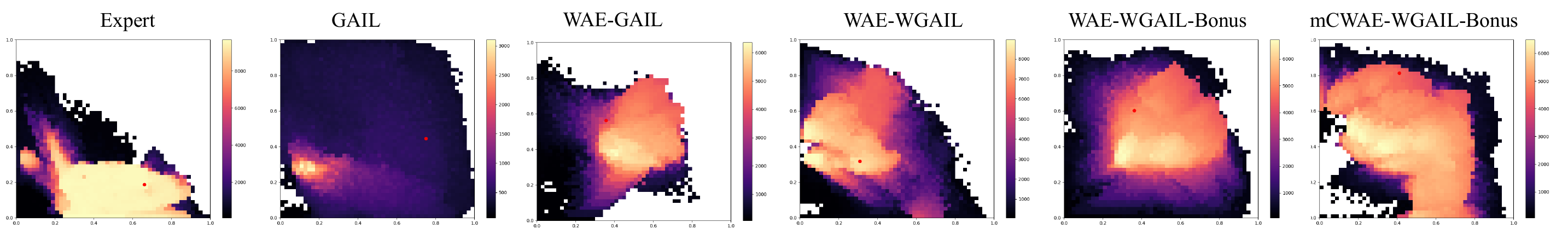}
    \vspace{-0.2cm}
    \caption{Visualization of the policy archive of Expert, GAIL, WAE-GAIL, WAE-WGAIL, WAE-WGAIL-Bonus and mCWAE-WGAIL-Bonus on Humanoid. The color indicates the cumulative rewards of best performing policy in the archive cells.}
    \label{fig:heatmap}
\end{figure*}
\begin{figure*}[htbp]
    \includegraphics[width=1.0\textwidth]{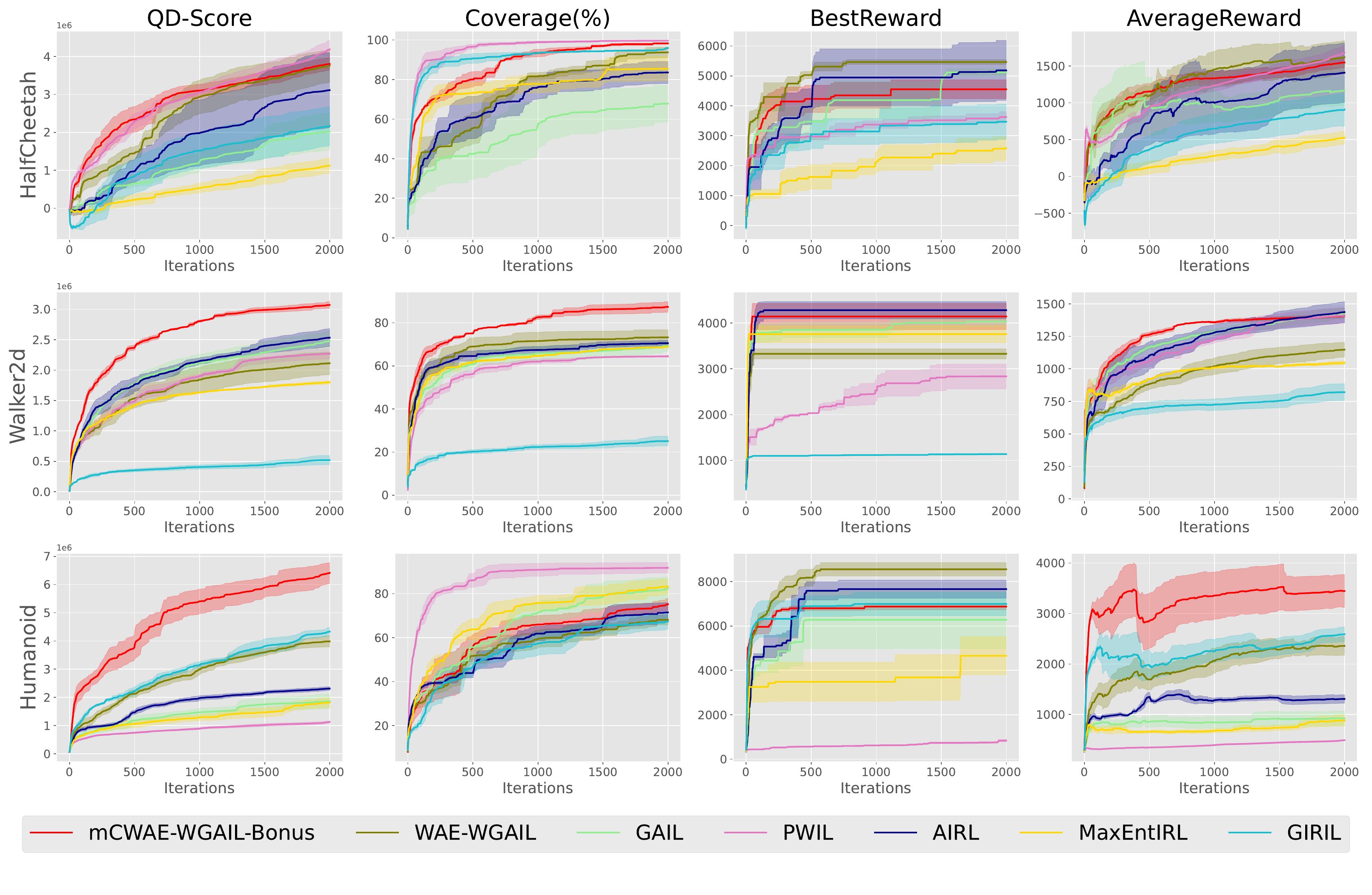}
    \vspace{-1.0cm}
    \caption{Learning curve comparison of our mCWAE-WGAIL-Bonus against WAE-WGAIL and state-of-the-art IL methods. The curves (and shaded areas) represent the means (and standard deviations) of the algorithms. Columns indicate the metric (QD-Score, Coverage, Best Reward, and Average Reward) while the rows represent the different benchmarks (Halfcheetah, Walker2d, and Humanoid).}
    \label{fig:main_results}
\end{figure*}

\section{Experiments}
\label{sect:expriments}

\subsection{Experiment Setup}
We evaluate our framework on three popular MuJoCo \citep{mujoco} environments: Halfcheetah, Humanoid, and Walker2d. The measure function is the number of times a leg contacts with the ground, divided by the trajectory length.
We implement all the methods based on the PPGA implementation, which utilizes the Brax simulator \citep{brax} with QDax wrappers for measure calculation \citep{QDax}. All experiments were conducted on A40 GPUs for three runs per task.

\subsection{Demonstrations}
\label{sect:exp_demo}
We use a policy archive obtained by PPGA to generate expert demonstrations. To follow real-world scenario with limited demonstrations, we first sample the top 500 high-performance elites from the archive as candidate pool and then select 4 policies from these candidates such that they are as diverse as possible. We generate 1 demonstration for each selected policy, resulting in the final 4 demonstrations for QDIL. This process results in 4 diverse demonstrations (episodes) per environment. Figure \ref{fig:top500elites} visualizes the demonstrators in the policy archive, and Table \ref{tab:top500elites} in Appendix \ref{sect:demo_details} provides the statistical properties of the demonstrations.

\subsection{Performance}
To validate the effectiveness of WQIL with Single-Step Archive Exploration, we compare our mCVAE-WAIL-Bonus against state-of-the-art IL methods within QDIL framework. These methods includes our WAE-WGAIL (Algorithm \ref{alg:WAE-WGAIL}, Appendix \ref{sect:algorithms}), GAIL \cite{ho2016generative}, PWIL \cite{PWIL}, AIRL \cite{AIRL}, MaxEntIRL \cite{max_ent} and GIRIL \cite{giril}. We apply a gradient penalty to all the methods that require the online update of the reward model. Appendix \ref{hyperpara} summarizes the hyperparameters. 

We compare the performance under four QD metrics \cite{PPGA}: 1) \textbf{QD-Score}, the sum of scores of all nonempty cells in the archive. QD-score is the most important metric in QDIL as it aligns with the objective of QDIL as in equation (\ref{obj}); 2) \textbf{Coverage}, the percentage of nonempty cells, indicating the algorithm's ability to discover diverse behaviors; 3) \textbf{Best Reward}, the highest score found by the algorithm; and 4) \textbf{Average Reward}, the mean score of all nonempty cells, reflecting the ability to discover both diverse and high-performing policies. We use the true reward functions to calculate these metrics.

The heatmaps in Figure \ref{fig:heatmap} visualize the performance across the policy archive for Expert, GAIL, WAE-GAIL, WAE-WGAIL, WAE-WGAIL-Bonus and mCWAE-WGAIL-Bonus on Humanoid. Even using true rewards, the expert explores less than half of the archive in the Humanoid environment. The high-performance policies are more restricted, covering around a quarter of the full archive. GAIL nearly explores the full archive space, however, most of the learned policies are low-performance (in dark color). Due to the training instability and behavior-overfitting issues, GAIL only learns high-performance policies within a small region of the behavior space. WAE-GAIL learns more high-performance policies than GAIL since it maintains better stability than GAIL. However, the explored region is still limited. By applying latent Wasserstein adversarial training, WAE-WGAIL further enlarges the explored archive region and coverage of high-performance policies. WAE-WGAIL-Bonus updates the archive using the rewards learned by WAE-WGAIL and the estimated single-step archive exploration bonus. The explored region of WAE-WGAIL-Bonus is clearly larger than WAE-WGAIL, which we attribute to the exploration bonus yielding improved coverage of the measure space. Thanks to the measure conditioning, the explored archive space of mCWAE-WGAIL-Bonus is slightly smaller but is more focused in the sense that it yields the largest set of high-performing policies among the five QDIL methods.

Figure \ref{fig:main_results} shows the learning curve comparison of QD performance for different QDIL methods. Among the four metrics, QD-Score is the most important because it reflects both high diversity and high quality of performance. In HalfCheetah, our method mCWAE-WGAIL-Bonus achieves one of the highest QD-Score with much better sample efficiency than the other counterparts (i.e., achieving the same QD-Score with much fewer iterations). In Walker2d and Humanoid environments, our method achieves a significantly higher QD-Score than the other methods.
PWIL achieves a slightly higher QD-Score than our method in HalfCheetah, however, it performs the worst in Humanoid. MaxEntIRL performs worse than GAIL, while AIRL outperforms GAIL slightly. Pretraining from limited demonstrations, GIRIL performs similar to GAIL in HalfCheetah and similar to WAE-WGAIL in Humanoid but much worse in Walker2d. 
For Coverage, our mCWAE-WGAIL-Bonus is the best in HalfCheetah and Walker2d. Although our coverage in Humanoid is not the best, we significantly outperform the baselines in terms of AverageReward. 
The quantitative results are summarized in Table \ref{tab:main_results} of Appendix \ref{sect:perf_table}.

\subsection{Ablations}
\label{sect:exp_ablation}
We conduct two ablations to study the effects of each component of our method mCWAE-WGAIL-Bonus. The first ablation in Section \ref{sect:ablation_wasserstein} studies the effects of latent Wasserstein adversarial training in the QDIL framework. On the other hand, Section \ref{sect:ablation_bonus_cond} focuses on studying the effects of single-step archive exploration bonus and measure conditioning.

\subsubsection{Ablation on Latent Wasserstein Adversarial Training}
\label{sect:ablation_wasserstein}
To study the effect of latent Wasserstein adversarial training, we additionally introduce mCWAE-GAIL-Bonus, which reinforces WAE-GAIL with measure conditioning and a single-step archive exploration bonus. 
Table \ref{tab:ablation_latent_wgail} in Appendix \ref{sect:ablation_curves} compares the QD performance of our mCWAE-WGAIL-Bonus with mCWAE-GAIL-Bonus and WAE-WGAIL. As we can see, applying Wasserstein adversarial training improves the QD-Score of WAE-GAIL and mCWAE-GAIL-Bonus by 27.5\% and 23.7\% respectively on HalfCheetah. In Walker2d, the latent Wasserserstein adversarial training enhances the QD-Score of WAE-GAIL and mCWAE-GAIL-Bonus by 74.3\% and 62.4\%, which is significant. Moreover in Humanoid, mCWAE-WGAIL-Bonus achieves 2x QD-Score of mCWAE-GAIL-Bonus, impressively outperforming the expert (i.e., PPGA-trueReward) by 12\%. The learning curves are shown in Figure \ref{fig:ablation_latent_wgail} in Appendix \ref{sect:ablation_curves}.

\subsubsection{Ablations on Single-Step Archive Exploration and Measure Conditioning}
\label{sect:ablation_bonus_cond}
Table \ref{tab:ablations_components} in of Appendix \ref{sect:ablation_curves} compares the effects of single-step archive exploration bonus and measure conditioning. Measure conditioning improves the QD-Score for WAE-WGAIL-Bonus by 42.8\%. In Walker2d, the contributions of measure conditioning to WAE-WGAL and WAE-WGAIL-Bonus are 12.3\% and 20.3\% respectively. In Humanoid, mCWAE-WGAIL-Bonus outperforms WAE-WGAIL-Bonus by 28.7\%. 
Now, we study the effect of the single-step archive exploration bonus. The bonus does not show much improvement on HalfCheetah. Comparing mCWAE-WGAIL-Bonus and mCWAE-WGAIL, we found that the contributions of bonus on the QD-Score improvement are 29.5\% in Walker2d and 80\% in Humanoid. Comparing WAE-WGAIL-Bonus and WAE-WGAIL, the contributions of the bonus on the QD-Score improvement are 20.8\% in Walker2d and 24.8\% in Humanoid. 
We illustrate the corresponding learning curves of this ablation study in Figure \ref{fig:ablation_bonus_cond} of Appendix \ref{sect:ablation_curves}.

\section{Conclusion}
\label{sect:conclusion}
This paper focuses on addressing two key issues of adversarial QDIL, i.e., the training instability of adversarial training and the behavior-overfitting in the design of the reward function. We propose WQDIL and a measure-conditioned reward function with Single-Step Archive Exploration to address these two issues respectively. Through two thorough ablation studies, we confirm the significance of each component (latent Wasserstein training, measure conditioning, and single-step exploration bonus) and find that latent Wasserstein adversarial training contributes most significantly to the QD-Score. The collaboration between these components contributes to the superior QD performance of our method on three continuous control tasks. Our method enables the agent to stably learn diverse and high-quality behaviors from limited demonstrations.



\begin{acks}
This project is supported by the National Research Foundation, Singapore and Infocomm Media Development Authority under its Trust Tech Funding Initiative. Any opinions, findings and conclusions or recommendations expressed in this material are those of the authors and do not reflect the views of National Research Foundation, Singapore and Infocomm Media Development Authority. Zhenglin Wan contributed to this publication whilst serving his internship with A*STAR CFAR under the Foreign Student Attachment scheme.
\end{acks}


\clearpage
\bibliographystyle{ACM-Reference-Format} 
\bibliography{ref}


\begin{thebibliography}{54}


\ifx \showCODEN    \undefined \def \showCODEN     #1{\unskip}     \fi
\ifx \showDOI      \undefined \def \showDOI       #1{#1}\fi
\ifx \showISBNx    \undefined \def \showISBNx     #1{\unskip}     \fi
\ifx \showISBNxiii \undefined \def \showISBNxiii  #1{\unskip}     \fi
\ifx \showISSN     \undefined \def \showISSN      #1{\unskip}     \fi
\ifx \showLCCN     \undefined \def \showLCCN      #1{\unskip}     \fi
\ifx \shownote     \undefined \def \shownote      #1{#1}          \fi
\ifx \showarticletitle \undefined \def \showarticletitle #1{#1}   \fi
\ifx \showURL      \undefined \def \showURL       {\relax}        \fi
\providecommand\bibfield[2]{#2}
\providecommand\bibinfo[2]{#2}
\providecommand\natexlab[1]{#1}
\providecommand\showeprint[2][]{arXiv:#2}

\bibitem[\protect\citeauthoryear{Abbeel and Ng}{Abbeel and Ng}{2004}]%
        {IRL}
\bibfield{author}{\bibinfo{person}{Pieter Abbeel} {and} \bibinfo{person}{Andrew~Y Ng}.} \bibinfo{year}{2004}\natexlab{}.
\newblock \showarticletitle{Apprenticeship learning via inverse reinforcement learning}. In \bibinfo{booktitle}{\emph{Proceedings of the twenty-first International Conference on Machine Learning}}.
\newblock


\bibitem[\protect\citeauthoryear{Arjovsky, Chintala, and Bottou}{Arjovsky et~al\mbox{.}}{2017}]%
        {arjovsky2017wasserstein}
\bibfield{author}{\bibinfo{person}{Martin Arjovsky}, \bibinfo{person}{Soumith Chintala}, {and} \bibinfo{person}{L{\'e}on Bottou}.} \bibinfo{year}{2017}\natexlab{}.
\newblock \showarticletitle{Wasserstein generative adversarial networks}. In \bibinfo{booktitle}{\emph{International conference on machine learning}}. PMLR, \bibinfo{pages}{214--223}.
\newblock


\bibitem[\protect\citeauthoryear{Auer, Cesa-Bianchi, and Fischer}{Auer et~al\mbox{.}}{2002}]%
        {Auer2002}
\bibfield{author}{\bibinfo{person}{Peter Auer}, \bibinfo{person}{Nicol{\`{o}} Cesa-Bianchi}, {and} \bibinfo{person}{Paul Fischer}.} \bibinfo{year}{2002}\natexlab{}.
\newblock \showarticletitle{{Finite-time analysis of the multiarmed bandit problem}}.
\newblock \bibinfo{journal}{\emph{Machine Learning}}  \bibinfo{volume}{47} (\bibinfo{year}{2002}), \bibinfo{pages}{235--256}.
\newblock
\showISBNx{0885-6125}
\showISSN{08856125}
\urldef\tempurl%
\url{https://doi.org/10.1023/A:1013689704352}
\showDOI{\tempurl}


\bibitem[\protect\citeauthoryear{Batra, Tjanaka, Fontaine, Petrenko, Nikolaidis, and Sukhatme}{Batra et~al\mbox{.}}{2023a}]%
        {batra2023proximal}
\bibfield{author}{\bibinfo{person}{Sumeet Batra}, \bibinfo{person}{Bryon Tjanaka}, \bibinfo{person}{Matthew~C Fontaine}, \bibinfo{person}{Aleksei Petrenko}, \bibinfo{person}{Stefanos Nikolaidis}, {and} \bibinfo{person}{Gaurav Sukhatme}.} \bibinfo{year}{2023}\natexlab{a}.
\newblock \showarticletitle{Proximal Policy Gradient Arborescence for Quality Diversity Reinforcement Learning}.
\newblock \bibinfo{journal}{\emph{arXiv preprint arXiv:2305.13795}} (\bibinfo{year}{2023}).
\newblock


\bibitem[\protect\citeauthoryear{Batra, Tjanaka, Fontaine, Petrenko, Nikolaidis, and Sukhatme}{Batra et~al\mbox{.}}{2023b}]%
        {PPGA}
\bibfield{author}{\bibinfo{person}{Sumeet Batra}, \bibinfo{person}{Bryon Tjanaka}, \bibinfo{person}{Matthew~C Fontaine}, \bibinfo{person}{Aleksei Petrenko}, \bibinfo{person}{Stefanos Nikolaidis}, {and} \bibinfo{person}{Gaurav Sukhatme}.} \bibinfo{year}{2023}\natexlab{b}.
\newblock \showarticletitle{Proximal policy gradient arborescence for quality diversity reinforcement learning}.
\newblock \bibinfo{journal}{\emph{arXiv preprint arXiv:2305.13795}} (\bibinfo{year}{2023}).
\newblock


\bibitem[\protect\citeauthoryear{Bojarski}{Bojarski}{2016}]%
        {driving}
\bibfield{author}{\bibinfo{person}{Mariusz Bojarski}.} \bibinfo{year}{2016}\natexlab{}.
\newblock \showarticletitle{End to end learning for self-driving cars}.
\newblock \bibinfo{journal}{\emph{arXiv preprint arXiv:1604.07316}} (\bibinfo{year}{2016}).
\newblock


\bibitem[\protect\citeauthoryear{Bousquet, Gelly, Tolstikhin, Simon-Gabriel, and Sch{\"{o}}lkopf}{Bousquet et~al\mbox{.}}{2017}]%
        {BGT+17}
\bibfield{author}{\bibinfo{person}{O. Bousquet}, \bibinfo{person}{S. Gelly}, \bibinfo{person}{I. Tolstikhin}, \bibinfo{person}{C.~J. Simon-Gabriel}, {and} \bibinfo{person}{B. Sch{\"{o}}lkopf}.} \bibinfo{year}{2017}\natexlab{}.
\newblock \bibinfo{title}{From optimal transport to generative modeling: the {VEGAN} cookbook}.
\newblock
\newblock
\showeprint{arXiv:1705.07642}


\bibitem[\protect\citeauthoryear{Brafman and Tennenholtz}{Brafman and Tennenholtz}{2002}]%
        {Brafman2002}
\bibfield{author}{\bibinfo{person}{Ronen~I Brafman} {and} \bibinfo{person}{Moshe Tennenholtz}.} \bibinfo{year}{2002}\natexlab{}.
\newblock \showarticletitle{{R-MAX -- A General Polynomial Time Algorithm for Near-Optimal Reinforcement Learning}}.
\newblock \bibinfo{journal}{\emph{Journal of Machine Learning Research}}  \bibinfo{volume}{3} (\bibinfo{year}{2002}), \bibinfo{pages}{213--231}.
\newblock
\showISSN{15324435}
\urldef\tempurl%
\url{https://doi.org/10.1162/153244303765208377}
\showDOI{\tempurl}


\bibitem[\protect\citeauthoryear{Burda, Edwards, Pathak, Storkey, Darrell, and Efros}{Burda et~al\mbox{.}}{2019}]%
        {burda2018large}
\bibfield{author}{\bibinfo{person}{Yuri Burda}, \bibinfo{person}{Harri Edwards}, \bibinfo{person}{Deepak Pathak}, \bibinfo{person}{Amos Storkey}, \bibinfo{person}{Trevor Darrell}, {and} \bibinfo{person}{Alexei~A Efros}.} \bibinfo{year}{2019}\natexlab{}.
\newblock \showarticletitle{Large-scale study of curiosity-driven learning}. In \bibinfo{booktitle}{\emph{International Conference on Learning Representations}}.
\newblock


\bibitem[\protect\citeauthoryear{Chatzilygeroudis, Cully, Vassiliades, and Mouret}{Chatzilygeroudis et~al\mbox{.}}{2021}]%
        {qd_definition}
\bibfield{author}{\bibinfo{person}{Konstantinos Chatzilygeroudis}, \bibinfo{person}{Antoine Cully}, \bibinfo{person}{Vassilis Vassiliades}, {and} \bibinfo{person}{Jean-Baptiste Mouret}.} \bibinfo{year}{2021}\natexlab{}.
\newblock \showarticletitle{Quality-diversity optimization: a novel branch of stochastic optimization}.
\newblock In \bibinfo{booktitle}{\emph{Black Box Optimization, Machine Learning, and No-Free Lunch Theorems}}. \bibinfo{publisher}{Springer}, \bibinfo{pages}{109--135}.
\newblock


\bibitem[\protect\citeauthoryear{Cideron, Pierrot, Perrin, Beguir, and Sigaud}{Cideron et~al\mbox{.}}{2020}]%
        {QD-robot}
\bibfield{author}{\bibinfo{person}{Geoffrey Cideron}, \bibinfo{person}{Thomas Pierrot}, \bibinfo{person}{Nicolas Perrin}, \bibinfo{person}{Karim Beguir}, {and} \bibinfo{person}{Olivier Sigaud}.} \bibinfo{year}{2020}\natexlab{}.
\newblock \showarticletitle{Qd-rl: Efficient mixing of quality and diversity in reinforcement learning}.
\newblock \bibinfo{journal}{\emph{arXiv preprint arXiv:2006.08505}} (\bibinfo{year}{2020}), \bibinfo{pages}{28--73}.
\newblock


\bibitem[\protect\citeauthoryear{Dadashi, Hussenot, Geist, and Pietquin}{Dadashi et~al\mbox{.}}{2021}]%
        {PWIL}
\bibfield{author}{\bibinfo{person}{Robert Dadashi}, \bibinfo{person}{L{\'e}onard Hussenot}, \bibinfo{person}{Matthieu Geist}, {and} \bibinfo{person}{Olivier Pietquin}.} \bibinfo{year}{2021}\natexlab{}.
\newblock \showarticletitle{Primal wasserstein imitation learning}. In \bibinfo{booktitle}{\emph{International Conference on Learning Representations}}.
\newblock


\bibitem[\protect\citeauthoryear{Finn, Levine, and Abbeel}{Finn et~al\mbox{.}}{2016}]%
        {max_ent_2}
\bibfield{author}{\bibinfo{person}{Chelsea Finn}, \bibinfo{person}{Sergey Levine}, {and} \bibinfo{person}{Pieter Abbeel}.} \bibinfo{year}{2016}\natexlab{}.
\newblock \showarticletitle{Guided cost learning: Deep inverse optimal control via policy optimization}. In \bibinfo{booktitle}{\emph{International Conference on Machine Learning}}. PMLR, \bibinfo{pages}{49--58}.
\newblock


\bibitem[\protect\citeauthoryear{Fontaine and Nikolaidis}{Fontaine and Nikolaidis}{2021}]%
        {DQD}
\bibfield{author}{\bibinfo{person}{Matthew Fontaine} {and} \bibinfo{person}{Stefanos Nikolaidis}.} \bibinfo{year}{2021}\natexlab{}.
\newblock \showarticletitle{Differentiable quality diversity}.
\newblock \bibinfo{journal}{\emph{Advances in Neural Information Processing Systems}}  \bibinfo{volume}{34} (\bibinfo{year}{2021}), \bibinfo{pages}{10040--10052}.
\newblock


\bibitem[\protect\citeauthoryear{Fontaine and Nikolaidis}{Fontaine and Nikolaidis}{2023}]%
        {CMA-MAEGA}
\bibfield{author}{\bibinfo{person}{Matthew Fontaine} {and} \bibinfo{person}{Stefanos Nikolaidis}.} \bibinfo{year}{2023}\natexlab{}.
\newblock \showarticletitle{Covariance matrix adaptation map-annealing}. In \bibinfo{booktitle}{\emph{Proceedings of the Genetic and Evolutionary Computation Conference}}. \bibinfo{pages}{456--465}.
\newblock


\bibitem[\protect\citeauthoryear{Fontaine, Togelius, Nikolaidis, and Hoover}{Fontaine et~al\mbox{.}}{2020}]%
        {CMA-ME}
\bibfield{author}{\bibinfo{person}{Matthew~C Fontaine}, \bibinfo{person}{Julian Togelius}, \bibinfo{person}{Stefanos Nikolaidis}, {and} \bibinfo{person}{Amy~K Hoover}.} \bibinfo{year}{2020}\natexlab{}.
\newblock \showarticletitle{Covariance matrix adaptation for the rapid illumination of behavior space}. In \bibinfo{booktitle}{\emph{Proceedings of the 2020 genetic and evolutionary computation conference}}. \bibinfo{pages}{94--102}.
\newblock


\bibitem[\protect\citeauthoryear{Freeman, Frey, Raichuk, Girgin, Mordatch, and Bachem}{Freeman et~al\mbox{.}}{2021}]%
        {brax}
\bibfield{author}{\bibinfo{person}{C~Daniel Freeman}, \bibinfo{person}{Erik Frey}, \bibinfo{person}{Anton Raichuk}, \bibinfo{person}{Sertan Girgin}, \bibinfo{person}{Igor Mordatch}, {and} \bibinfo{person}{Olivier Bachem}.} \bibinfo{year}{2021}\natexlab{}.
\newblock \showarticletitle{Brax-a differentiable physics engine for large scale rigid body simulation, 2021}.
\newblock \bibinfo{journal}{\emph{URL http://github. com/google/brax}}  \bibinfo{volume}{6} (\bibinfo{year}{2021}).
\newblock


\bibitem[\protect\citeauthoryear{Fu, Luo, and Levine}{Fu et~al\mbox{.}}{2017}]%
        {AIRL}
\bibfield{author}{\bibinfo{person}{Justin Fu}, \bibinfo{person}{Katie Luo}, {and} \bibinfo{person}{Sergey Levine}.} \bibinfo{year}{2017}\natexlab{}.
\newblock \showarticletitle{Learning robust rewards with adversarial inverse reinforcement learning}.
\newblock \bibinfo{journal}{\emph{arXiv preprint arXiv:1710.11248}} (\bibinfo{year}{2017}).
\newblock


\bibitem[\protect\citeauthoryear{Goodfellow, Pouget-Abadie, Mirza, Xu, Warde-Farley, Ozair, Courville, and Bengio}{Goodfellow et~al\mbox{.}}{2014}]%
        {goodfellow2014generative}
\bibfield{author}{\bibinfo{person}{Ian Goodfellow}, \bibinfo{person}{Jean Pouget-Abadie}, \bibinfo{person}{Mehdi Mirza}, \bibinfo{person}{Bing Xu}, \bibinfo{person}{David Warde-Farley}, \bibinfo{person}{Sherjil Ozair}, \bibinfo{person}{Aaron Courville}, {and} \bibinfo{person}{Yoshua Bengio}.} \bibinfo{year}{2014}\natexlab{}.
\newblock \showarticletitle{Generative adversarial nets}. In \bibinfo{booktitle}{\emph{Advances in Neural Information Processing Systems}}. \bibinfo{pages}{2672--2680}.
\newblock


\bibitem[\protect\citeauthoryear{Gulrajani, Ahmed, Arjovsky, Dumoulin, and Courville}{Gulrajani et~al\mbox{.}}{2017}]%
        {gulrajani2017improved}
\bibfield{author}{\bibinfo{person}{Ishaan Gulrajani}, \bibinfo{person}{Faruk Ahmed}, \bibinfo{person}{Martin Arjovsky}, \bibinfo{person}{Vincent Dumoulin}, {and} \bibinfo{person}{Aaron~C Courville}.} \bibinfo{year}{2017}\natexlab{}.
\newblock \showarticletitle{Improved training of wasserstein gans}.
\newblock \bibinfo{journal}{\emph{Advances in neural information processing systems}}  \bibinfo{volume}{30} (\bibinfo{year}{2017}).
\newblock


\bibitem[\protect\citeauthoryear{Hansen and Ostermeier}{Hansen and Ostermeier}{2001}]%
        {cma-es}
\bibfield{author}{\bibinfo{person}{Nikolaus Hansen} {and} \bibinfo{person}{Andreas Ostermeier}.} \bibinfo{year}{2001}\natexlab{}.
\newblock \showarticletitle{Completely derandomized self-adaptation in evolution strategies}.
\newblock \bibinfo{journal}{\emph{Evolutionary computation}} \bibinfo{volume}{9}, \bibinfo{number}{2} (\bibinfo{year}{2001}), \bibinfo{pages}{159--195}.
\newblock


\bibitem[\protect\citeauthoryear{Henaff, Raileanu, Jiang, and Rockt{\"{a}}schel}{Henaff et~al\mbox{.}}{2022}]%
        {Henaff2022}
\bibfield{author}{\bibinfo{person}{Mikael Henaff}, \bibinfo{person}{Roberta Raileanu}, \bibinfo{person}{Minqi Jiang}, {and} \bibinfo{person}{Tim Rockt{\"{a}}schel}.} \bibinfo{year}{2022}\natexlab{}.
\newblock \showarticletitle{{Exploration via Elliptical Episodic Bonuses}}. In \bibinfo{booktitle}{\emph{Advances in Neural Information Processing Systems (NeurIPS 2022)}}.
\newblock


\bibitem[\protect\citeauthoryear{Ho and Ermon}{Ho and Ermon}{2016}]%
        {ho2016generative}
\bibfield{author}{\bibinfo{person}{Jonathan Ho} {and} \bibinfo{person}{Stefano Ermon}.} \bibinfo{year}{2016}\natexlab{}.
\newblock \showarticletitle{Generative adversarial imitation learning}.
\newblock \bibinfo{journal}{\emph{Advances in neural information processing systems}}  \bibinfo{volume}{29} (\bibinfo{year}{2016}).
\newblock


\bibitem[\protect\citeauthoryear{Jaksch, Ortner, and Auer}{Jaksch et~al\mbox{.}}{2010}]%
        {Jaksch2010}
\bibfield{author}{\bibinfo{person}{Thomas Jaksch}, \bibinfo{person}{Ronald Ortner}, {and} \bibinfo{person}{Peter Auer}.} \bibinfo{year}{2010}\natexlab{}.
\newblock \showarticletitle{{Near-optimal regret bounds for reinforcement learning}}.
\newblock \bibinfo{journal}{\emph{Journal of Machine Learning Research}}  \bibinfo{volume}{11} (\bibinfo{year}{2010}), \bibinfo{pages}{1563--1600}.
\newblock
\showISSN{15324435}


\bibitem[\protect\citeauthoryear{Kearns and Singh}{Kearns and Singh}{2002}]%
        {Kearns2002a}
\bibfield{author}{\bibinfo{person}{Michael Kearns} {and} \bibinfo{person}{Satinder Singh}.} \bibinfo{year}{2002}\natexlab{}.
\newblock \showarticletitle{{Near-optimal reinforcement learning in polynomial time}}.
\newblock \bibinfo{journal}{\emph{Machine Learning}} \bibinfo{volume}{49}, \bibinfo{number}{2-3} (\bibinfo{year}{2002}), \bibinfo{pages}{209--232}.
\newblock
\showISSN{08856125}
\urldef\tempurl%
\url{https://doi.org/10.1023/A:1017984413808}
\showDOI{\tempurl}


\bibitem[\protect\citeauthoryear{Kingma and Welling}{Kingma and Welling}{2014}]%
        {KW14}
\bibfield{author}{\bibinfo{person}{D.~P. Kingma} {and} \bibinfo{person}{M. Welling}.} \bibinfo{year}{2014}\natexlab{}.
\newblock \showarticletitle{Auto-Encoding Variational {B}ayes}. In \bibinfo{booktitle}{\emph{International Conference on Learning Representations}}.
\newblock


\bibitem[\protect\citeauthoryear{Kolouri, Pope, Martin, and Rohde}{Kolouri et~al\mbox{.}}{2019}]%
        {kolouri2018sliced}
\bibfield{author}{\bibinfo{person}{Soheil Kolouri}, \bibinfo{person}{Phillip~E. Pope}, \bibinfo{person}{Charles~E. Martin}, {and} \bibinfo{person}{Gustavo~K. Rohde}.} \bibinfo{year}{2019}\natexlab{}.
\newblock \showarticletitle{Sliced Wasserstein Auto-Encoders}. In \bibinfo{booktitle}{\emph{International Conference on Learning Representations}}.
\newblock
\urldef\tempurl%
\url{https://openreview.net/forum?id=H1xaJn05FQ}
\showURL{%
\tempurl}


\bibitem[\protect\citeauthoryear{Lim, Allard, Grillotti, and Cully}{Lim et~al\mbox{.}}{2022}]%
        {QDax}
\bibfield{author}{\bibinfo{person}{Bryan Lim}, \bibinfo{person}{Maxime Allard}, \bibinfo{person}{Luca Grillotti}, {and} \bibinfo{person}{Antoine Cully}.} \bibinfo{year}{2022}\natexlab{}.
\newblock \showarticletitle{Accelerated quality-diversity for robotics through massive parallelism}. In \bibinfo{booktitle}{\emph{ICLR Workshop on Agent Learning in Open-Endedness}}.
\newblock


\bibitem[\protect\citeauthoryear{Luo, Pearce, Chen, Chen, and Zhu}{Luo et~al\mbox{.}}{2024}]%
        {luo2024c}
\bibfield{author}{\bibinfo{person}{Tianjiao Luo}, \bibinfo{person}{Tim Pearce}, \bibinfo{person}{Huayu Chen}, \bibinfo{person}{Jianfei Chen}, {and} \bibinfo{person}{Jun Zhu}.} \bibinfo{year}{2024}\natexlab{}.
\newblock \showarticletitle{C-GAIL: Stabilizing Generative Adversarial Imitation Learning with Control Theory}.
\newblock \bibinfo{journal}{\emph{arXiv preprint arXiv:2402.16349}} (\bibinfo{year}{2024}).
\newblock


\bibitem[\protect\citeauthoryear{Machado, Bellemare, and Bowling}{Machado et~al\mbox{.}}{2020}]%
        {Machado2020}
\bibfield{author}{\bibinfo{person}{Marlos~C. Machado}, \bibinfo{person}{Marc~G. Bellemare}, {and} \bibinfo{person}{Michael Bowling}.} \bibinfo{year}{2020}\natexlab{}.
\newblock \showarticletitle{{Count-based exploration with the successor representation}}.
\newblock \bibinfo{journal}{\emph{AAAI 2020 - 34th AAAI Conference on Artificial Intelligence}} (\bibinfo{year}{2020}), \bibinfo{pages}{5123--5133}.
\newblock
\showISBNx{9781577358350}
\showISSN{2159-5399}
\showeprint[arxiv]{1807.11622}


\bibitem[\protect\citeauthoryear{Mescheder, Geiger, and Nowozin}{Mescheder et~al\mbox{.}}{2018}]%
        {mescheder2018training}
\bibfield{author}{\bibinfo{person}{Lars Mescheder}, \bibinfo{person}{Andreas Geiger}, {and} \bibinfo{person}{Sebastian Nowozin}.} \bibinfo{year}{2018}\natexlab{}.
\newblock \showarticletitle{Which training methods for GANs do actually converge?}. In \bibinfo{booktitle}{\emph{International conference on machine learning}}. PMLR, \bibinfo{pages}{3481--3490}.
\newblock


\bibitem[\protect\citeauthoryear{Mouret and Clune}{Mouret and Clune}{2015}]%
        {map-elites}
\bibfield{author}{\bibinfo{person}{Jean-Baptiste Mouret} {and} \bibinfo{person}{Jeff Clune}.} \bibinfo{year}{2015}\natexlab{}.
\newblock \showarticletitle{Illuminating search spaces by mapping elites}.
\newblock \bibinfo{journal}{\emph{arXiv preprint arXiv:1504.04909}} (\bibinfo{year}{2015}).
\newblock


\bibitem[\protect\citeauthoryear{Pathak, Agrawal, Efros, and Darrell}{Pathak et~al\mbox{.}}{2017}]%
        {pathak2017curiosity}
\bibfield{author}{\bibinfo{person}{Deepak Pathak}, \bibinfo{person}{Pulkit Agrawal}, \bibinfo{person}{Alexei~A Efros}, {and} \bibinfo{person}{Trevor Darrell}.} \bibinfo{year}{2017}\natexlab{}.
\newblock \showarticletitle{Curiosity-driven exploration by self-supervised prediction}. In \bibinfo{booktitle}{\emph{International Conference on Machine Learning}}. PMLR, \bibinfo{pages}{2778--2787}.
\newblock


\bibitem[\protect\citeauthoryear{Petzka, Fischer, and Lukovnicov}{Petzka et~al\mbox{.}}{2017}]%
        {petzka2017regularization}
\bibfield{author}{\bibinfo{person}{Henning Petzka}, \bibinfo{person}{Asja Fischer}, {and} \bibinfo{person}{Denis Lukovnicov}.} \bibinfo{year}{2017}\natexlab{}.
\newblock \showarticletitle{On the regularization of wasserstein gans}.
\newblock \bibinfo{journal}{\emph{arXiv preprint arXiv:1709.08894}} (\bibinfo{year}{2017}).
\newblock


\bibitem[\protect\citeauthoryear{Pugh, Soros, and Stanley}{Pugh et~al\mbox{.}}{2016}]%
        {QD_def}
\bibfield{author}{\bibinfo{person}{Justin~K Pugh}, \bibinfo{person}{Lisa~B Soros}, {and} \bibinfo{person}{Kenneth~O Stanley}.} \bibinfo{year}{2016}\natexlab{}.
\newblock \showarticletitle{Quality diversity: A new frontier for evolutionary computation}.
\newblock \bibinfo{journal}{\emph{Frontiers in Robotics and AI}}  \bibinfo{volume}{3} (\bibinfo{year}{2016}), \bibinfo{pages}{202845}.
\newblock


\bibitem[\protect\citeauthoryear{Qi}{Qi}{2020}]%
        {qi2020loss}
\bibfield{author}{\bibinfo{person}{Guo-Jun Qi}.} \bibinfo{year}{2020}\natexlab{}.
\newblock \showarticletitle{Loss-sensitive generative adversarial networks on lipschitz densities}.
\newblock \bibinfo{journal}{\emph{International Journal of Computer Vision}} \bibinfo{volume}{128}, \bibinfo{number}{5} (\bibinfo{year}{2020}), \bibinfo{pages}{1118--1140}.
\newblock


\bibitem[\protect\citeauthoryear{Ross, Gordon, and Bagnell}{Ross et~al\mbox{.}}{2011}]%
        {BC_limitation}
\bibfield{author}{\bibinfo{person}{St{\'e}phane Ross}, \bibinfo{person}{Geoffrey Gordon}, {and} \bibinfo{person}{Drew Bagnell}.} \bibinfo{year}{2011}\natexlab{}.
\newblock \showarticletitle{A reduction of imitation learning and structured prediction to no-regret online learning}. In \bibinfo{booktitle}{\emph{Proceedings of the fourteenth international conference on artificial intelligence and statistics}}. JMLR Workshop and Conference Proceedings, \bibinfo{pages}{627--635}.
\newblock


\bibitem[\protect\citeauthoryear{Ross, Melik-Barkhudarov, Shankar, Wendel, Dey, Bagnell, and Hebert}{Ross et~al\mbox{.}}{2013}]%
        {drone}
\bibfield{author}{\bibinfo{person}{St{\'e}phane Ross}, \bibinfo{person}{Narek Melik-Barkhudarov}, \bibinfo{person}{Kumar~Shaurya Shankar}, \bibinfo{person}{Andreas Wendel}, \bibinfo{person}{Debadeepta Dey}, \bibinfo{person}{J~Andrew Bagnell}, {and} \bibinfo{person}{Martial Hebert}.} \bibinfo{year}{2013}\natexlab{}.
\newblock \showarticletitle{Learning monocular reactive uav control in cluttered natural environments}. In \bibinfo{booktitle}{\emph{2013 IEEE international conference on robotics and automation}}. IEEE, \bibinfo{pages}{1765--1772}.
\newblock


\bibitem[\protect\citeauthoryear{Roth, Lucchi, Nowozin, and Hofmann}{Roth et~al\mbox{.}}{2017}]%
        {roth2017stabilizing}
\bibfield{author}{\bibinfo{person}{Kevin Roth}, \bibinfo{person}{Aurelien Lucchi}, \bibinfo{person}{Sebastian Nowozin}, {and} \bibinfo{person}{Thomas Hofmann}.} \bibinfo{year}{2017}\natexlab{}.
\newblock \showarticletitle{Stabilizing training of generative adversarial networks through regularization}.
\newblock \bibinfo{journal}{\emph{Advances in neural information processing systems}}  \bibinfo{volume}{30} (\bibinfo{year}{2017}).
\newblock


\bibitem[\protect\citeauthoryear{Schulman, Wolski, Dhariwal, Radford, and Klimov}{Schulman et~al\mbox{.}}{2017}]%
        {PPO}
\bibfield{author}{\bibinfo{person}{John Schulman}, \bibinfo{person}{Filip Wolski}, \bibinfo{person}{Prafulla Dhariwal}, \bibinfo{person}{Alec Radford}, {and} \bibinfo{person}{Oleg Klimov}.} \bibinfo{year}{2017}\natexlab{}.
\newblock \showarticletitle{Proximal policy optimization algorithms}.
\newblock \bibinfo{journal}{\emph{arXiv preprint arXiv:1707.06347}} (\bibinfo{year}{2017}).
\newblock


\bibitem[\protect\citeauthoryear{Seo, Chen, Shin, Lee, Abbeel, and Lee}{Seo et~al\mbox{.}}{2021}]%
        {Seo2021}
\bibfield{author}{\bibinfo{person}{Younggyo Seo}, \bibinfo{person}{Lili Chen}, \bibinfo{person}{Jinwoo Shin}, \bibinfo{person}{Honglak Lee}, \bibinfo{person}{Pieter Abbeel}, {and} \bibinfo{person}{Kimin Lee}.} \bibinfo{year}{2021}\natexlab{}.
\newblock \showarticletitle{{State Entropy Maximization with Random Encoders for Efficient Exploration}}.
\newblock \bibinfo{journal}{\emph{Proceedings of Machine Learning Research}}  \bibinfo{volume}{139} (\bibinfo{year}{2021}), \bibinfo{pages}{9443--9454}.
\newblock


\bibitem[\protect\citeauthoryear{Sohn, Lee, and Yan}{Sohn et~al\mbox{.}}{2015}]%
        {sohn2015learning}
\bibfield{author}{\bibinfo{person}{Kihyuk Sohn}, \bibinfo{person}{Honglak Lee}, {and} \bibinfo{person}{Xinchen Yan}.} \bibinfo{year}{2015}\natexlab{}.
\newblock \showarticletitle{Learning structured output representation using deep conditional generative models}. In \bibinfo{booktitle}{\emph{Advances in Neural Information Processing Systems (NeurIPS 2015)}}.
\newblock


\bibitem[\protect\citeauthoryear{Tang, Houthooft, Foote, Stooke, Chen, Duan, Schulman, {De Turck}, and Abbeel}{Tang et~al\mbox{.}}{2017}]%
        {Tang2017}
\bibfield{author}{\bibinfo{person}{Haoran Tang}, \bibinfo{person}{Rein Houthooft}, \bibinfo{person}{Davis Foote}, \bibinfo{person}{Adam Stooke}, \bibinfo{person}{Xi Chen}, \bibinfo{person}{Yan Duan}, \bibinfo{person}{John Schulman}, \bibinfo{person}{Filip {De Turck}}, {and} \bibinfo{person}{Pieter Abbeel}.} \bibinfo{year}{2017}\natexlab{}.
\newblock \showarticletitle{{\#Exploration: A study of count-based exploration for deep reinforcement learning}}. In \bibinfo{booktitle}{\emph{Advances in Neural Information Processing Systems (NeurIPS 2017)}}. \bibinfo{pages}{2754--2763}.
\newblock


\bibitem[\protect\citeauthoryear{Thanh-Tung, Tran, and Venkatesh}{Thanh-Tung et~al\mbox{.}}{2019}]%
        {thanh2019improving}
\bibfield{author}{\bibinfo{person}{Hoang Thanh-Tung}, \bibinfo{person}{Truyen Tran}, {and} \bibinfo{person}{Svetha Venkatesh}.} \bibinfo{year}{2019}\natexlab{}.
\newblock \showarticletitle{Improving generalization and stability of generative adversarial networks}.
\newblock \bibinfo{journal}{\emph{arXiv preprint arXiv:1902.03984}} (\bibinfo{year}{2019}).
\newblock


\bibitem[\protect\citeauthoryear{Tjanaka, Fontaine, Togelius, and Nikolaidis}{Tjanaka et~al\mbox{.}}{2022}]%
        {tjanaka2022approximating}
\bibfield{author}{\bibinfo{person}{Bryon Tjanaka}, \bibinfo{person}{Matthew~C Fontaine}, \bibinfo{person}{Julian Togelius}, {and} \bibinfo{person}{Stefanos Nikolaidis}.} \bibinfo{year}{2022}\natexlab{}.
\newblock \showarticletitle{Approximating gradients for differentiable quality diversity in reinforcement learning}. In \bibinfo{booktitle}{\emph{Proceedings of the Genetic and Evolutionary Computation Conference}}. \bibinfo{pages}{1102--1111}.
\newblock


\bibitem[\protect\citeauthoryear{Todorov, Erez, and Tassa}{Todorov et~al\mbox{.}}{2012}]%
        {mujoco}
\bibfield{author}{\bibinfo{person}{Emanuel Todorov}, \bibinfo{person}{Tom Erez}, {and} \bibinfo{person}{Yuval Tassa}.} \bibinfo{year}{2012}\natexlab{}.
\newblock \showarticletitle{Mujoco: A physics engine for model-based control}. In \bibinfo{booktitle}{\emph{2012 IEEE/RSJ international conference on intelligent robots and systems}}. IEEE, \bibinfo{pages}{5026--5033}.
\newblock


\bibitem[\protect\citeauthoryear{Tolstikhin, Bousquet, Gelly, and Schoelkopf}{Tolstikhin et~al\mbox{.}}{2018}]%
        {tolstikhin2017wasserstein}
\bibfield{author}{\bibinfo{person}{Ilya Tolstikhin}, \bibinfo{person}{Olivier Bousquet}, \bibinfo{person}{Sylvain Gelly}, {and} \bibinfo{person}{Bernhard Schoelkopf}.} \bibinfo{year}{2018}\natexlab{}.
\newblock \showarticletitle{Wasserstein auto-encoders}. In \bibinfo{booktitle}{\emph{International Conference on Learning Representations}}.
\newblock


\bibitem[\protect\citeauthoryear{Villani}{Villani}{2003}]%
        {V03}
\bibfield{author}{\bibinfo{person}{C. Villani}.} \bibinfo{year}{2003}\natexlab{}.
\newblock \bibinfo{booktitle}{\emph{Topics in Optimal Transportation}}.
\newblock \bibinfo{publisher}{AMS Graduate Studies in Mathematics}.
\newblock


\bibitem[\protect\citeauthoryear{Wan, Yu, Bossens, Lyu, Guo, Fan, and Tsang}{Wan et~al\mbox{.}}{2024}]%
        {wan2024qualitydiversityimitationlearning}
\bibfield{author}{\bibinfo{person}{Zhenglin Wan}, \bibinfo{person}{Xingrui Yu}, \bibinfo{person}{David~Mark Bossens}, \bibinfo{person}{Yueming Lyu}, \bibinfo{person}{Qing Guo}, \bibinfo{person}{Flint~Xiaofeng Fan}, {and} \bibinfo{person}{Ivor Tsang}.} \bibinfo{year}{2024}\natexlab{}.
\newblock \showarticletitle{Quality Diversity Imitation Learning}.
\newblock  (\bibinfo{year}{2024}).
\newblock
\showeprint[arxiv]{2410.06151}~[cs.LG]
\urldef\tempurl%
\url{https://arxiv.org/abs/2410.06151}
\showURL{%
\tempurl}


\bibitem[\protect\citeauthoryear{Wulfmeier, Ondruska, and Posner}{Wulfmeier et~al\mbox{.}}{2015}]%
        {max_ent_1}
\bibfield{author}{\bibinfo{person}{Markus Wulfmeier}, \bibinfo{person}{Peter Ondruska}, {and} \bibinfo{person}{Ingmar Posner}.} \bibinfo{year}{2015}\natexlab{}.
\newblock \showarticletitle{Maximum entropy deep inverse reinforcement learning}.
\newblock \bibinfo{journal}{\emph{arXiv preprint arXiv:1507.04888}} (\bibinfo{year}{2015}).
\newblock


\bibitem[\protect\citeauthoryear{Yu, Lyu, and Tsang}{Yu et~al\mbox{.}}{2020}]%
        {giril}
\bibfield{author}{\bibinfo{person}{Xingrui Yu}, \bibinfo{person}{Yueming Lyu}, {and} \bibinfo{person}{Ivor Tsang}.} \bibinfo{year}{2020}\natexlab{}.
\newblock \showarticletitle{Intrinsic reward driven imitation learning via generative model}. In \bibinfo{booktitle}{\emph{International Conference on Machine Learning}}. PMLR, \bibinfo{pages}{10925--10935}.
\newblock


\bibitem[\protect\citeauthoryear{Zare, Kebria, Khosravi, and Nahavandi}{Zare et~al\mbox{.}}{2024}]%
        {IL_survey}
\bibfield{author}{\bibinfo{person}{Maryam Zare}, \bibinfo{person}{Parham~M Kebria}, \bibinfo{person}{Abbas Khosravi}, {and} \bibinfo{person}{Saeid Nahavandi}.} \bibinfo{year}{2024}\natexlab{}.
\newblock \showarticletitle{A survey of imitation learning: Algorithms, recent developments, and challenges}.
\newblock \bibinfo{journal}{\emph{IEEE Transactions on Cybernetics}} (\bibinfo{year}{2024}).
\newblock


\bibitem[\protect\citeauthoryear{Zhu, Wang, Merel, Rusu, Erez, Cabi, Tunyasuvunakool, Kram{\'a}r, Hadsell, de~Freitas, et~al\mbox{.}}{Zhu et~al\mbox{.}}{2018}]%
        {manipulation}
\bibfield{author}{\bibinfo{person}{Yuke Zhu}, \bibinfo{person}{Ziyu Wang}, \bibinfo{person}{Josh Merel}, \bibinfo{person}{Andrei Rusu}, \bibinfo{person}{Tom Erez}, \bibinfo{person}{Serkan Cabi}, \bibinfo{person}{Saran Tunyasuvunakool}, \bibinfo{person}{J{\'a}nos Kram{\'a}r}, \bibinfo{person}{Raia Hadsell}, \bibinfo{person}{Nando de Freitas}, {et~al\mbox{.}}} \bibinfo{year}{2018}\natexlab{}.
\newblock \showarticletitle{Reinforcement and imitation learning for diverse visuomotor skills}.
\newblock \bibinfo{journal}{\emph{arXiv preprint arXiv:1802.09564}} (\bibinfo{year}{2018}).
\newblock


\bibitem[\protect\citeauthoryear{Ziebart, Maas, Bagnell, Dey, et~al\mbox{.}}{Ziebart et~al\mbox{.}}{2008}]%
        {max_ent}
\bibfield{author}{\bibinfo{person}{Brian~D Ziebart}, \bibinfo{person}{Andrew~L Maas}, \bibinfo{person}{J~Andrew Bagnell}, \bibinfo{person}{Anind~K Dey}, {et~al\mbox{.}}} \bibinfo{year}{2008}\natexlab{}.
\newblock \showarticletitle{Maximum entropy inverse reinforcement learning.}. In \bibinfo{booktitle}{\emph{AAAI Conference on Artificial Intelligence}}, Vol.~\bibinfo{volume}{8}. Chicago, IL, USA, \bibinfo{pages}{1433--1438}.
\newblock


\end{thebibliography}



\appendix
\onecolumn 


\section{Algorithms}
\label{sect:algorithms}

\begin{algorithm}[!h]\captionsetup{labelfont={sc,bf}}
\captionof{algorithm}{Wasserstein Auto-Encoder with GAN-based penalty (WAE-GAN).}
\label{alg:WAE-GAN}
\renewcommand{\algorithmicensure}{\textbf{Output:}}
  \begin{algorithmic}[1]
  \State \textbf{Require:} Regularization coefficient $\lambda>0$.
  \State Initialize the parameters of the encoder $Q_\phi$, \\decoder $G_\psi$, and latent discriminator $D$.
  \While{$(\phi, \psi)$ not converged}
  
  \State Sample $\{x_1,\dots,x_n\}$ from the training set          
  
  \State Sample $\{z_1,\dots,z_n\}$ from the prior $P_z$
  
  \State Sample $z_i$ from $Q_\phi(z|x_i)$ for $i=1,\dots,n$
  
  \State Update $D$ by ascending:
  \[\frac{\lambda}{n} \sum_{i=1}^n \log D(z_i) + \log \bigl(1-D(z_i)\bigr)\]

\vspace{.05cm}
  \State Update $Q_\phi$ and $G_\psi$ by descending:
  \[\frac{1}{n}\sum_{i=1}^n f_c\bigl(x_i, G_\psi(z_i)\bigr) - \lambda \cdot \log D(z_i)\]
  \EndWhile
  \end{algorithmic}
\end{algorithm}

\begin{algorithm}[htbp]\captionsetup{labelfont={sc,bf}}
\captionof{algorithm}{WAE-GAIL reward model updates.}
\label{alg:WAE-GAIL}
\renewcommand{\algorithmicensure}{\textbf{Output:}}
  \begin{algorithmic}[1]
  \State \textbf{Input:} Expert demonstrations $\mathcal{D}$, Current policy $\pi_\theta$, Regularization coefficient $\lambda>0$.
  \State Initialize the parameters of the encoder $Q_\phi$, \\decoder $G_\psi$, and latent discriminator $D$.
  
  \vspace{0.5cm} 
  \State \textbf{Method: Update reward model in Algorithm \ref{alg:wqdil} (step 17)}
  \State \texttt{def update}($\mathcal{D}, \pi_\theta$): 
  
  \State \quad Sample expert trajectories $\{(s^e_1, a^e_1), \cdots , (s^e_n, a^e_n) \}$ from the expert demonstration $\mathcal{D}$ 
  
  \State \quad Sample trajectories $\{(s^\pi_1, a^\pi_1), \cdots , (s^\pi_n, a^\pi_n) \}$ from the policy $\pi_\theta$ 
  
  \State \quad Sample $\{z_1,\dots,z_n\}$ from the prior $P_z$

  \State \quad Sample $z^e_i$ from $Q_\phi(z|s^e_i,a^e_i)$ for $i=1,\dots,n$
  
  \State \quad Sample $z^\pi_i$ from $Q_\phi(z|s^\pi_i,a^\pi_i)$ for $i=1,\dots,n$
  
  \State \quad Update $D$ by ascending:
  \[\frac{\lambda}{n} \sum_{i=1}^n \log D(z_i) + \log D(z^e_i) + \log \bigl(1-D(z^\pi_i)\bigr)\]

  \vspace{.05cm}
  \State \quad Update $Q_\phi$ and $G_\psi$ by descending:
  \[\frac{1}{n}\sum_{i=1}^n \Big[ \Big(f_c\bigl((s^e_i, a^e_i), G_\psi(z^e_i)\bigr) + f_c\bigl((s^\pi_i, a^\pi_i), G_\psi(z^\pi_i)\bigr)\Big)
  - \lambda \cdot \Big( \log D(z^e_i) + \log D(z^\pi_i) \Big)\Big]
  \] 
  \State \quad Repeat until the model converges or the number of epochs is reached
  \State \quad \textbf{return} Updated $Q_\phi$, $G_\psi$ and $D$
  \end{algorithmic}
\end{algorithm}

\begin{algorithm}[htbp]\captionsetup{labelfont={sc,bf}}
\captionof{algorithm}{WAE-WGAIL reward model updates.}
\label{alg:WAE-WGAIL}
\renewcommand{\algorithmicensure}{\textbf{Output:}}
  \begin{algorithmic}[1]
  \State \textbf{Input:} Expert demonstrations $\mathcal{D}$, Current policy $\pi_\theta$, Regularization coefficient $\lambda>0$.
  \State Initialize the parameters of the encoder $Q_\phi$, \\decoder $G_\psi$, and latent Wasserstein discriminator $D_\mathcal{W}$.

  \vspace{0.5cm}
  \State \textbf{Method: Update reward model in Algorithm \ref{alg:wqdil} (step 17)}
  \State \texttt{def update}($\mathcal{D}, \pi_\theta$): 
  \State \quad Sample expert trajectories $\{(s^e_1, a^e_1), \cdots , (s^e_n, a^e_n) \}$ from the expert demonstration $\mathcal{D}$ 
  
  \State \quad Sample trajectories $\{(s^\pi_1, a^\pi_1), \cdots , (s^\pi_n, a^\pi_n) \}$ from the policy $\pi_\theta$ 

  \State \quad Sample $z^e_i$ from $Q_\phi(z|s^e_i,a^e_i)$ for $i=1,\dots,n$
  
  \State \quad Sample $z^\pi_i$ from $Q_\phi(z|s^\pi_i,a^\pi_i)$ for $i=1,\dots,n$
  
  \State \quad Update $D_\mathcal{W}$ by ascending:
  \[\frac{\lambda}{n} \sum_{i=1}^n  D_\mathcal{W}(z^e_i) -D_\mathcal{W}(z^\pi_i)\]

\vspace{.05cm}
  \State \quad Update $Q_\phi$ and $G_\psi$ by descending:
  \[\frac{1}{n}\sum_{i=1}^n \Big(f_c\bigl((s^e_i, a^e_i), G_\psi(z^e_i)\bigr) + f_c\bigl((s^\pi_i, a^\pi_i), G_\psi(z^\pi_i)\bigr)\Big)
  \] 
  \State \quad Repeat until the model converges or the number of epochs is reached
  \State \quad \textbf{return} Updated $Q_\phi$, $G_\psi$ and $D_\mathcal{W}$
  \end{algorithmic}
\end{algorithm}

\begin{algorithm}[htbp]\captionsetup{labelfont={sc,bf}}
\captionof{algorithm}{mCWAE-WGAIL reward model updates.}
\label{alg:mCWAE-WGAIL}
\renewcommand{\algorithmicensure}{\textbf{Output:}}
  \begin{algorithmic}[1]
  \State \textbf{Input:} Expert demonstrations $\mathcal{D}$, Current policy $\pi_\theta$, Regularization coefficient $\lambda>0$.
  \State Initialize the parameters of the encoder $Q_\phi$, \\decoder $G_\psi$, and latent Wasserstein discriminator $D_\mathcal{W}$.

  \State \textbf{Method: Update reward model in Algorithm \ref{alg:wqdil} (step 17)}
  \State \texttt{def update}($\mathcal{D}, \pi_\theta$): 
  
  \State \quad Sample expert data $\{(s^e_1, a^e_1, \delta(s)^e_1), \cdots , (s^e_n, a^e_n, \delta(s)^e_n) \}$ from the expert demonstration $\mathcal{D}$ 
  
  \State \quad Sample trajectories $\{(s^\pi_1, a^\pi_1, \delta(s)^\pi_1), \cdots , (s^\pi_n, a^\pi_n, \delta(s)^\pi_n) \}$ from the policy $\pi_\theta$ 

  \State \quad Sample $z^e_i$ from $Q_\phi(z|s^e_i,a^e_i,\delta(s)^e_i)$ for $i=1,\dots,n$
  
  \State \quad Sample $z^\pi_i$ from $Q_\phi(z|s^\pi_i,a^\pi_i,\delta(s)^\pi_i)$ for $i=1,\dots,n$
  
  \State \quad Update $D_\mathcal{W}$ by ascending:
  \[\frac{\lambda}{n} \sum_{i=1}^n  D_\mathcal{W}(z^e_i) -D_\mathcal{W}(z^\pi_i)\]

\vspace{.05cm}
  \State \quad Update $Q_\phi$ and $G_\psi$ by descending:
  \[\frac{1}{n}\sum_{i=1}^n \Big(f_c\bigl((s^e_i, a^e_i, \delta(s)^e_i), G_\psi(z^e_i)\bigr) + f_c\bigl((s^\pi_i, a^\pi_i, \delta(s)^\pi_i), G_\psi(z^\pi_i)\bigr)\Big)
  \] 
  \State \quad Repeat until the model converges or the number of epochs is reached
  \State \quad \textbf{return} Updated $Q_\phi$, $G_\psi$ and $D_\mathcal{W}$
  \end{algorithmic}
\end{algorithm}

\newpage

\section{Hyperparameter Setting}
\label{hyperpara}

\subsection{Hyperparameters for PPGA}
Table \ref{tab:hyperpara_ppga} summarizes the list of hyperparameters for PPGA policy updates.
\begin{table}[!h]
    \caption{List of relevant hyperparameters for PPGA shared across all environments.}
    \centering
    \begin{tabular}{c|c}
    \toprule
        Hyperparameter & Value \\
        \midrule
        Actor Network & [128, 128, Action Dim] \\ 
        Critic Network & [256, 256, 1] \\ 
        $N_1$ & 10 \\
        $N_2$ & 10 \\
        PPO Num Minibatches & 8 \\
        PPO Num Epochs & 4 \\
        Observation Normalization & True \\
        Reward Normalization & True \\
        Rollout Length & 128 \\
        Grid Size & 50 \\
        Env Batch Size & 3,000 \\
        Num iterations & 2,000 \\
        \bottomrule
    \end{tabular}
    \label{tab:hyperpara_ppga}
\end{table}

\subsection{Hyperparameters for IL}

Table \ref{tab:hyperpara_gail} summarizes the list of hyperparameters for AIRL, GAIL, measure-conditioned GAIL, and MConbo-GAIL. 

\begin{table}[!h]
    \caption{List of relevant hyperparameters for GAIL and WAEs shared across all environments.}
    \centering
    \begin{tabular}{c|c}
    \toprule
        Hyperparameter & Value \\
        \midrule
        Discriminator & [100, 100, 1] \\ 
        Learning Rate & 3e-4 \\
        Discriminator Num Epochs & 1 \\
        Regularization coefficient $\lambda$ & 1  \\
        \bottomrule
    \end{tabular}
    \label{tab:hyperpara_gail}
\end{table}

Table \ref{tab:hyperpara_girl} summarizes the list of hyperparameters for GIRIL. 
\begin{table}[htbp]
    \caption{List of relevant hyperparameters for GIRIL shared across all environments.}
    \centering
    \begin{tabular}{c|c}
    \toprule
        Hyperparameter & Value \\
        \midrule
        Encoder & [100, 100, Action Dim] \\ 
        Decoder & [100, 100, Observation Dim] \\
        Learning Rate & 3e-4 \\
        Batch Size & 32 \\
        Num Pretrain Epochs & 10,000 \\ 
        \bottomrule
    \end{tabular}
    \label{tab:hyperpara_girl}
\end{table}

\newpage
\section{Demonstration Details}
\label{sect:demo_details}

\begin{table}[htbp]
    \centering
    \caption{Demonstration statistics on the three environments.}
    \scalebox{1.0}{
    \begin{tabular}{c|c|c|c|c|c|c}
    \toprule
      Tasks & Demo num & Attributes & min & max & mean & std \\ \midrule
        \multirow{2}{*}{HalfCheetah} & \multirow{2}{*}{4} & Length & 1000 & 1000 & 1000.0 & 0.0   \\
           & & Return & 3766.0 & 8405.4 & 5721.3 & 1927.6 \\ 
          
           \midrule
        \multirow{2}{*}{Walker2d} & \multirow{2}{*}{4} & Length & 356.0 & 1000.0 & 625.8 & 254.4 \\
           & & Return & 1147.9 & 3721.8 & 2372.3 & 1123.7  \\ 
           
           \midrule
        \multirow{2}{*}{Humanoid} & \multirow{2}{*}{4} & Length & 1000.0 & 1000.0 & 1000.0 & 0.0 \\
           & & Return & 7806.2 & 9722.6 & 8829.5 & 698.1\\ 
    \bottomrule
    \end{tabular}
    }
    \label{tab:top500elites}
\end{table}

\clearpage 

\section{Additional Experimental Results}
\label{sect:additional_results}

\subsection{Overall results: quantitative performance and Tukey HSD test}
\label{sect:perf_table}

\begin{table*}[htbp]
\centering
\scriptsize
\caption{QD performance comparison of our mCWAE-WGAIL-Bonus against WAE-WGAIL, GAIL, PWIL, AIRL, MaxEntIRL and GIRIL across three environments. Cov, Best, and Avg refer to Coverage, Best Reward, and Average Reward respectively.  Bold indicates the highest two metric scores.}
\resizebox{\textwidth}{!}{
\begin{tabular}{l c c c c c c c c c c c c}
\toprule
& \multicolumn{4}{c}{HalfCheetah} & \multicolumn{4}{c}{Walker2d} & \multicolumn{4}{c}{Humanoid} \\
\cmidrule(lr){2-5} \cmidrule(lr){6-9} \cmidrule(lr){10-13}
& QD-Score & Cov(\%) & Best & Avg & QD-Score & Cov(\%) & Best & Avg & QD-Score & Cov(\%) & Best & Avg \\
\midrule
PPGA-trueReward & $6.75 \times 10^6$ & 94.08 & 8,942 & 2,871 & $3.64 \times 10^6$ & 77.04 & 5,588 & 1,891 & $5.71 \times 10^6$ & 49.96 & 9,691 & 4,570 \\
\midrule
mCWAE-WGAIL-Bonus & $\mathbf{3.80 \times 10^6}\pm\mathbf{2.85\times10^5}$ & \textbf{98.28$\pm$ 0.42} & 4,553$\pm$547 & 1,547$\pm$118 & $\mathbf{3.07 \times 10^6}\pm\mathbf{9.75\times10^4}$ & \textbf{87.39}$\pm$\textbf{4.32} & \textbf{4,142$\pm$496} &  1,407$\pm$26 & $\mathbf{6.40 \times 10^6}\pm\mathbf{6.31\times10^5}$ & 75.05$\pm$4.82 & 6,875$\pm$254 & \textbf{3,447$\pm$562} \\
WAE-WGAIL  & $3.74 \times 10^6 \pm 7.55 \times10^5$ & 93.77 $\pm$4.67 & \textbf{5,463$\pm$136} & \textbf{1,620$\pm$395} & $2.11 \times 10^6 \pm 3.25 \times10^5$ & \textbf{73.20$\pm$6.24} & 3,327$\pm$192 & 1,148$\pm$99 & $3.98 \times 10^6\pm 3.38\times 10^5$ & 68.08$\pm$3.9 & \textbf{8,553$\pm$521} & 2,359$\pm$321 \\
PWIL & $\mathbf{4.17 \times 10^6}\pm\mathbf{4.07\times10^5}$ & \textbf{99.65$\pm$0.28} & 3,626$\pm$312 & \textbf{1,682$\pm$165} & $2.27 \times 10^6\pm1.49\times10^5$ & 64.45$\pm$0.66 & 2,835$\pm$470 & \textbf{1,410$\pm$106} & $1.12 \times 10^6\pm7.20\times10^4$ & \textbf{91.73$\pm$4.11} & 841$\pm$96 & 492$\pm$24 \\
AIRL & $3.11 \times 10^6\pm1.71\times10^6$ & 83.57$\pm$9.65 & \textbf{5,183$\pm$1,735} & 1,410$\pm$700 & $\mathbf{2.53 \times 10^6\pm2.59\times10^5}$ & 70.53$\pm$2.3 & \textbf{4,280$\pm$326} & \textbf{1,437$\pm$141} & $2.31 \times 10^6\pm1.15\times10^5$ & 71.47$\pm$7.59 & \textbf{7,661$\pm$705} & 1,308$\pm$144 \\
MaxEntIRL & $1.12 \times 10^6 \pm 3.54 \times 10^5$ & 85.48$\pm$10.65 & 2594$\pm$680 & 525$\pm$152 & $1.80 \times 10^6 \pm 5.13 \times 10^5$ & 68.83$\pm$0.73 & 3,756$\pm$318 & 1,046$\pm$29 & $1.82 \times 10^6 \pm 3.12 \times 10^5$ & \textbf{83.27$\pm$5.77} & 4,658$\pm$1489 & 882$\pm$190 \\
GIRIL & $2.17 \times 10^6 \pm 8.85 \times 10^5$ & 95.96$\pm$0.79 & 3,466$\pm$1018 & 909$\pm$377 & $0.52 \times 10^6  \pm 1.33 \times 10^5$ & 25.08$\pm$3.93 & 1,139$\pm$25 & 821$\pm$112 & $\mathbf{4.33 \times 10^6 \pm 2.58 \times 10^5}$ & 67.40$\pm$6.05 & 6,992$\pm$915 & \textbf{2,590$\pm$254} \\
\bottomrule
\end{tabular}
}
\label{tab:main_results}
\end{table*}

\begin{table}[h]
    \centering
    \caption{Tukey HSD test of the QD Score for the comparisons to baselines.  $+$ indicates a positive effect of mCWAE-WGAIL-Bonus.  The significant pairwise comparisons ($p$<=0.05) are boldfaced. }
    \begin{tabular}{cc|cc|cc|cc}
    \toprule
    group1  & group2  & \multicolumn{2}{c}{HalfCheetah} & \multicolumn{2}{c}{Walker2d} & \multicolumn{2}{c}{Humanoid} \\
    \midrule
     AIRL & mCWAE-WGAIL-Bonus & + & 0.9825 & + & 0.2094 & + & \textbf{0.0}\\
     GAIL & mCWAE-WGAIL-Bonus & + & 0.4375 & + & 0.1432 & + & \textbf{0.0}\\
    GIRIL & mCWAE-WGAIL-Bonus & + & 0.5333 & + & \textbf{0.0} & + & \textbf{0.0007}\\
MaxEntIRL & mCWAE-WGAIL-Bonus & + & 0.0928 & + & \textbf{0.0005} & + & \textbf{0.0}\\
     PWIL & mCWAE-WGAIL-Bonus & - & 0.9992 & + & \textbf{0.0241} & + & \textbf{0.0}\\
    \bottomrule
    \end{tabular}
    \label{tab:tukey_final_results}
\end{table}

\clearpage

\subsection{Ablation Study: quantitative performance, learning curves, and Tukey HSD test}
\label{sect:ablation_curves}

\begin{table*}[htbp]
\centering
\scriptsize
\caption{Ablation study results on the effect of latent Wasserstein adversarial training (-WGAIL). Bold indicates the highest metric score.}
\vspace{-0.3cm}
\resizebox{\textwidth}{!}{
\begin{tabular}{l c c c c c c c c c c c c}
\toprule
& \multicolumn{4}{c}{HalfCheetah} & \multicolumn{4}{c}{Walker2d} & \multicolumn{4}{c}{Humanoid} \\
\cmidrule(lr){2-5} \cmidrule(lr){6-9} \cmidrule(lr){10-13}
& QD-Score & Cov(\%) & Best & Avg & QD-Score & Cov(\%) & Best & Avg & QD-Score & Cov(\%) & Best & Avg \\
\midrule
PPGA-trueReward & $6.75 \times 10^6$ & 94.08 & 8,942 & 2,871 & $3.64 \times 10^6$ & 77.04 & 5,588 & 1,891 & $5.71 \times 10^6$ & 49.96 & 9,691 & 4,570 \\
\midrule
mCWAE-WGAIL-Bonus & $\mathbf{3.80 \times 10^6}\pm\mathbf{2.85\times10^5}$ & 98.28$\pm$ 0.42 & 4,553$\pm$547 & 1,547$\pm$118 & $\mathbf{3.07 \times 10^6}\pm\mathbf{9.75\times10^4}$ & \textbf{87.39}$\pm$\textbf{4.32} & \textbf{4,142$\pm$496} & \textbf{1,407$\pm$26} & $\mathbf{6.40 \times 10^6}\pm\mathbf{6.31\times10^5}$ & \textbf{75.05$\pm$4.82} & 6,875$\pm$254 & \textbf{3,447$\pm$562} \\
mCWAE-GAIL-Bonus  & $3.07 \times 10^7\pm4.03\times10^5$ & \textbf{99.32}$\pm$\textbf{0.64} & 3,350$\pm$570 & 1,235$\pm$158 & $1.89 \times 10^6\pm 2.11\times10^5$ & 56.88$\pm$2.39 & 1,860$\pm$310 & 1,337$\pm$178 & $3.26 \times 10^6\pm3.03\times10^5$ & 55.95$\pm$17.41 & 5,182$\pm$395 & 2,542$\pm$639 \\
WAE-WGAIL  & $3.74 \times 10^6\pm7.55\times10^5$ & 93.77 $\pm$4.67 & \textbf{5,463$\pm$136} & \textbf{1,620$\pm$395} & $2.11 \times 10^6\pm3.25\times10^5$ & 73.20$\pm$6.24 & 3,327$\pm$192 & 1,148$\pm$99 & $3.98 \times 10^6\pm3.38\times10^5$ & 68.08$\pm$3.9 & \textbf{8,553$\pm$521} & 2,359$\pm$321 \\
WAE-GAIL  & $2.91 \times 10^6\pm2.02\times10^6$ & 92.85$\pm$10.05 & 3,792$\pm$1,227 & 1,178$\pm$794 & $1.21 \times 10^6\pm3.98\times10^5$ & 43.59$\pm$13.15 & 1,357$\pm$147 & 1,097$\pm$58 & $4.01 \times 10^6\pm1.15\times10^6$ & 66.45$\pm$10.99 & 5,721$\pm$847 & 2,381$\pm$347 \\
\bottomrule
\end{tabular}
}
\label{tab:ablation_latent_wgail}
\end{table*}

\begin{table*}[htbp]
\centering
\scriptsize
\caption{Ablation study results on the effects of single-step archive exploration (-Bonus) and measure conditioning (mC). Bold indicates the highest metric score.}
\vspace{-0.3cm}
\resizebox{\textwidth}{!}{
\begin{tabular}{l c c c c c c c c c c c c}
\toprule
& \multicolumn{4}{c}{HalfCheetah} & \multicolumn{4}{c}{Walker2d} & \multicolumn{4}{c}{Humanoid} \\
\cmidrule(lr){2-5} \cmidrule(lr){6-9} \cmidrule(lr){10-13}
& QD-Score & Cov(\%) & Best & Avg & QD-Score & Cov(\%) & Best & Avg & QD-Score & Cov(\%) & Best & Avg \\
\midrule
PPGA-trueReward & $6.75 \times 10^6$ & 94.08 & 8,942 & 2,871 & $3.64 \times 10^6$ & 77.04 & 5,588 & 1,891 & $5.71 \times 10^6$ & 49.96 & 9,691 & 4,570 \\
\midrule
mCWAE-WGAIL-Bonus & $\mathbf{3.80 \times 10^6}\pm\mathbf{2.85\times10^5}$ & \textbf{98.28$\pm$ 0.42} & 4,553$\pm$547 & 1,547$\pm$118 & $\mathbf{3.07 \times 10^6}\pm\mathbf{9.75\times10^4}$ & \textbf{87.39}$\pm$\textbf{4.32} & \textbf{4,142$\pm$496} & \textbf{1,407$\pm$26} & $\mathbf{6.40 \times 10^6}\pm\mathbf{6.31\times10^5}$ & \textbf{75.05$\pm$4.82} & 6,875$\pm$254 & \textbf{3,447$\pm$562} \\
WAE-WGAIL-Bonus  & $2.66 \times 10^7\pm2.01\times10^6$ & 76.89$\pm$15.46 & 4,788$\pm$1,339 & 1,246$\pm$700 & $2.55 \times 10^6\pm1.28\times10^5$ & 73.93$\pm$2.58 & 3,310$\pm$114 & 1,381$\pm$99 & $4.97 \times 10^6\pm7.55\times10^5$ & 67.27$\pm$12.64 & \textbf{8,620$\pm$732} & 3,099$\pm$826 \\
mCWAE-WGAIL  & $3.79 \times 10^7\pm 1.89 \times 10^{6}$ & 87.76$\pm$12.95 & \textbf{5,728$\pm$1429} & \textbf{1,623$\pm$693} & $2.37 \times 10^6 \pm 9.3 \times 10^5$ & 71.61$\pm$2.0 & 3,611$\pm$524 & 1,326$\pm$73 & $3.54 \times 10^6 \pm 3.09 \times 10^5$ & 66.96$\pm$0.69 & 8,485$\pm$290 & 2,114$\pm$194 \\
WAE-WGAIL  & $3.74 \times 10^6 \pm 7.55 \times10^5$ & 93.77 $\pm$4.67 & 5,463$\pm$136 & 1,620$\pm$395 & $2.11 \times 10^6 \pm 3.2 \times10^5$ & 73.20$\pm$6.24 & 3,327$\pm$192 & 1,148$\pm$99 & $3.98 \times 10^6\pm 3.38\times 10^5$ & 68.08$\pm$3.9 & 8,553$\pm$521 & 2,359$\pm$321 \\
\bottomrule
\end{tabular}
}
\label{tab:ablations_components}
\end{table*}

\begin{figure*}[!h]
    \includegraphics[width=1.0\textwidth]{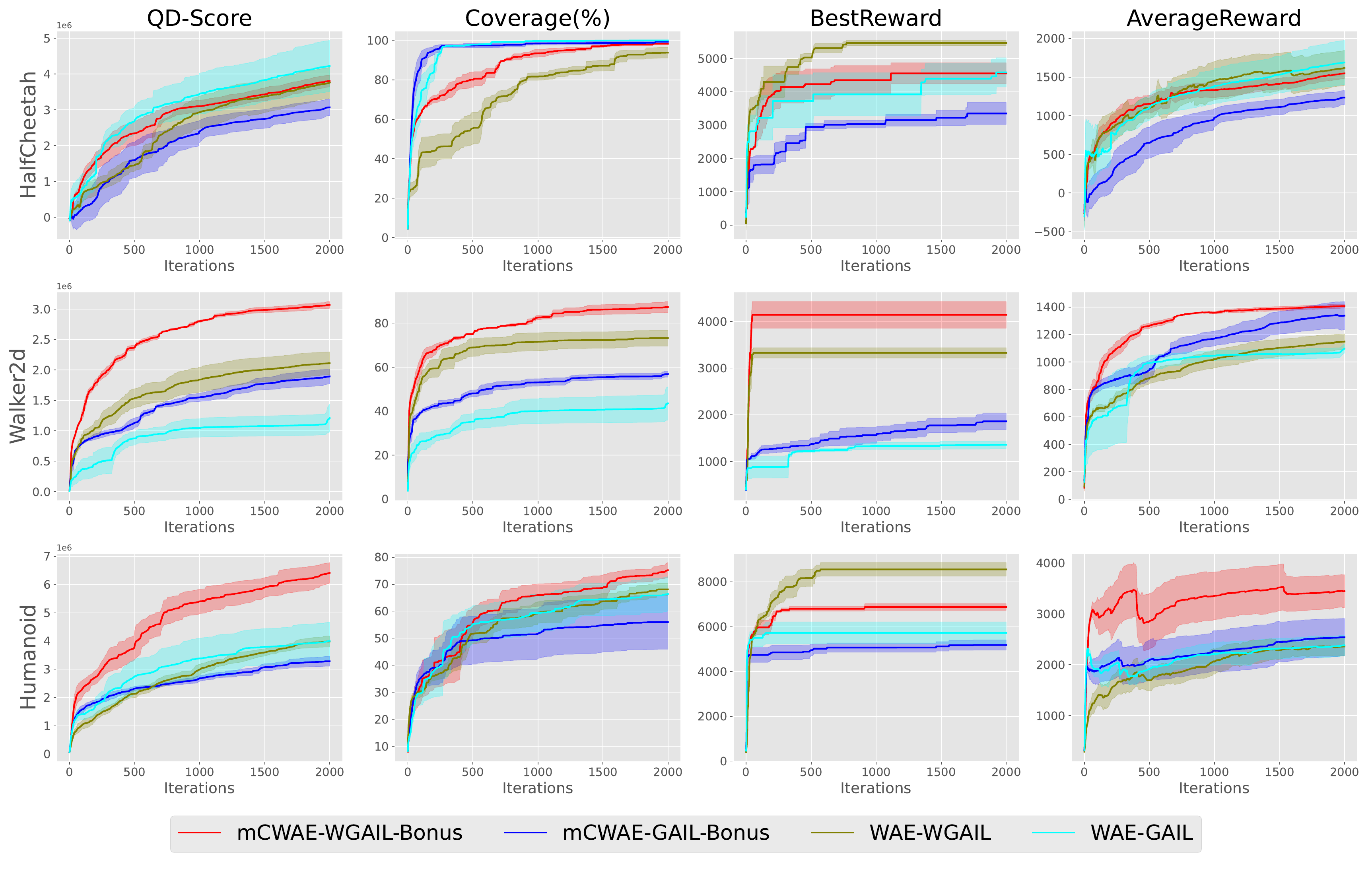}
    \caption{Learning curve comparison of our mCWAE-WGAIL-Bonus against WAE-GAIL-Bonus, WAE-WGAIL and WAE-GAIL. }
    \label{fig:ablation_latent_wgail}
\end{figure*}

\begin{table}[h]
    \centering
    \caption{Tukey HSD test of the QD Score for the ablation study on Wasserstein adversarial training. $+$ indicates a positive effect of mCWAE-WGAIL-Bonus. The significant pairwise comparisons ($p$<=0.05) are boldfaced. }
    \begin{tabular}{cc|cc|cc|cc}
    \toprule
    group1  & group2  & \multicolumn{2}{c}{HalfCheetah} & \multicolumn{2}{c}{Walker2d} & \multicolumn{2}{c}{Humanoid} \\
    \midrule
            GAIL & mCWAE-WGAIL-Bonus & + & 0.4844 & + & 0.3009 & + & \textbf{0.0003 }\\
        WAE-GAIL & mCWAE-WGAIL-Bonus & + & 0.9127 & + & \textbf{0.0005} & + & \textbf{0.027} \\
       WAE-WGAIL & mCWAE-WGAIL-Bonus & + & 1.0 & + & \textbf{0.0433} & + & \textbf{0.0256} \\
mCWAE-GAIL-Bonus & mCWAE-WGAIL-Bonus & + & 0.9533 &  + & \textbf{0.0133} & + &\textbf{0.0051} \\
    \bottomrule
    \end{tabular}
    \label{tab:tukey_ablation_latent_wgail_mC_bonus}
\end{table}

\begin{figure*}[!h]
    \includegraphics[width=1.0\textwidth]{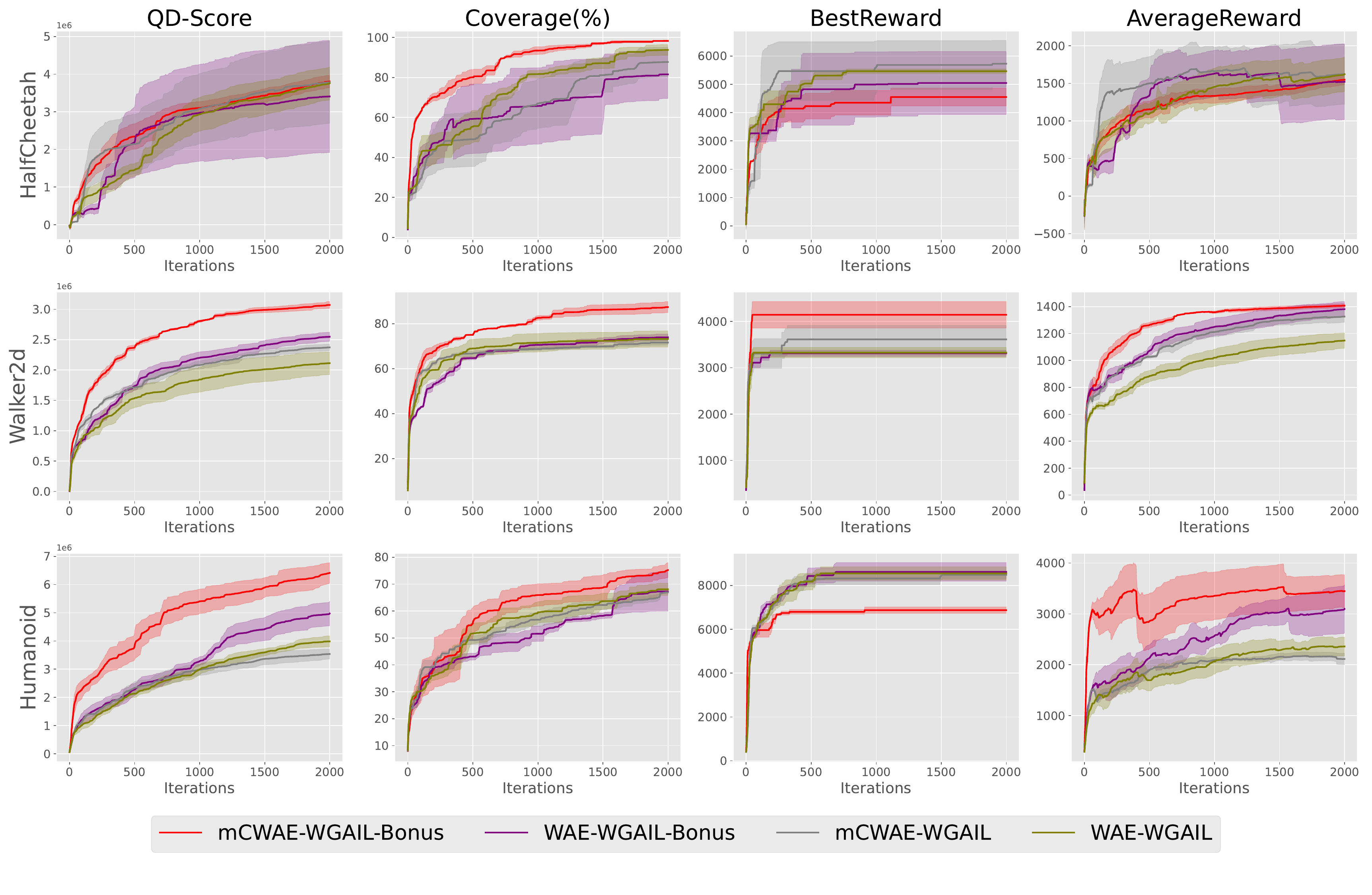}
    \caption{Learning curve comparison of our mCWAE-WGAIL-Bonus against WAE-WGAIL-Bonus, mCWAE-WGAIL and WAE-WGAIL. }
    \label{fig:ablation_bonus_cond}
\end{figure*}

\begin{table}[h]
    \centering
    \caption{Tukey HSD test of the QD Score for the ablation study on bonus and measure conditioning. $+$ indicates a positive effect of mCWAE-WGAIL-Bonus. The significant pairwise comparisons ($p$<=0.05) are boldfaced. }
    \begin{tabular}{cc|cc|cc|cc}
    \toprule
    group1  & group2  & \multicolumn{2}{c}{HalfCheetah} & \multicolumn{2}{c}{Walker2d} & \multicolumn{2}{c}{Humanoid} \\
    \midrule
      WAE-WGAIL & mCWAE-WGAIL-Bonus & + & 1.0 & + & \textbf{0.004} & + & \textbf{0.0088}\\
WAE-WGAIL-Bonus & mCWAE-WGAIL-Bonus & + & 0.8562 & + & 0.0886 & + & 0.1079 \\
mCWAE-WGAIL & mCWAE-WGAIL-Bonus & + & 1.0 & + & \textbf{0.0239 }& + & \textbf{0.0032 }\\
    \bottomrule
    \end{tabular}
    \label{tab:tukey_ablation_latent_wgail_bonus}
\end{table}

\end{document}